\documentclass[]{bytedance}
\usepackage[toc,page,header]{appendix}
\usepackage{minitoc}
\usepackage{algorithm}
\usepackage{algpseudocode}
\usepackage{booktabs}
\usepackage{multirow}
\usepackage{siunitx}
\usepackage{makecell}
\usepackage{mathtools}
\usepackage{tabularx,array}
\usepackage[table]{xcolor}
\usepackage{graphicx}
\usepackage{wrapfig}
\usepackage{xspace}
\usepackage{amsmath}
\usepackage{amsfonts}
\usepackage{epsfig}
\usepackage{amssymb}
\usepackage{etoolbox}
\usepackage{listings}
\usepackage{enumitem}
\usepackage{amsthm}

\usepackage{graphbox}
\usepackage{hyperref}

\title{\textsc{Seeing Fast and Slow}:\\Learning the Flow of Time in Videos}

\newcommand{\eg}{\textit{e.g.}}
\newcommand{\ie}{\textit{i.e.}}

\newcommand{\myparagraph}[1]{\noindent\textbf{#1}}

\usepackage{animate}
\newcommand{\dataset}{SloMo-44K\xspace}
\newcommand{\IfDefinedSwitch}[3]{%
  \ifdefined#1
    #2 
  \else
    #3 
  \fi
}
\newcommand{\embedVideo}{embed video}

\author[1,2]{Yen-Siang Wu}
\author[1]{Rundong Luo}
\author[1]{Jingsen Zhu}
\author[1]{Tao Tu}
\author[3]{Ali Farhadi}
\author[3]{\\Matthew Wallingford}
\author[2]{Yu-Chiang Frank Wang}
\author[1]{Steve Marschner}
\author[1]{Wei-Chiu Ma}

\affiliation[1]{Cornell University}
\affiliation[2]{National Taiwan University}
\affiliation[3]{University of Washington}

\abstract{
How can we tell whether a video has been sped up or slowed down? How can we generate videos at different speeds? Although videos have been central to modern computer vision research,  little attention has been paid to perceiving and controlling the passage of time. In this paper, we study time as a learnable visual concept and develop models for reasoning about and manipulating the flow of time in videos.

We first exploit the multimodal cues and temporal structure naturally present in videos to learn, in a self-supervised manner, to detect speed changes and estimate playback speed. We then show that these learned temporal reasoning models enable us to curate the largest slow-motion video dataset to date from noisy in-the-wild sources. Such slow-motion footage, typically filmed by high-speed cameras, contains substantially richer temporal detail than standard videos. Using this data, we further develop models capable of temporal control, including speed-conditioned video generation, which produces motion at specified playback speed, and temporal super-resolution, which tranforms low-FPS, blurry videos into high-FPS sequences with fine-grained temporal details.
Our findings highlight time as a manipulable, perceptual dimension in video learning, opening doors to temporally controllable video generation, temporal forensics detection, and potentially richer world-models that understand how events unfold over time.
}

\checkdata[Project Page]{\url{https://seeing-fast-and-slow.github.io/}}
\begin{document}
\maketitle

\begin{figure}[t]
    \centering
    \IfDefinedSwitch{\embedVideo}{
    \animategraphics[autoplay,loop,width=\linewidth, trim=0 0cm 0 0cm, clip]{30}{figures/teaser/}{001}{155}
    }{
        \includegraphics[width=1.\linewidth]{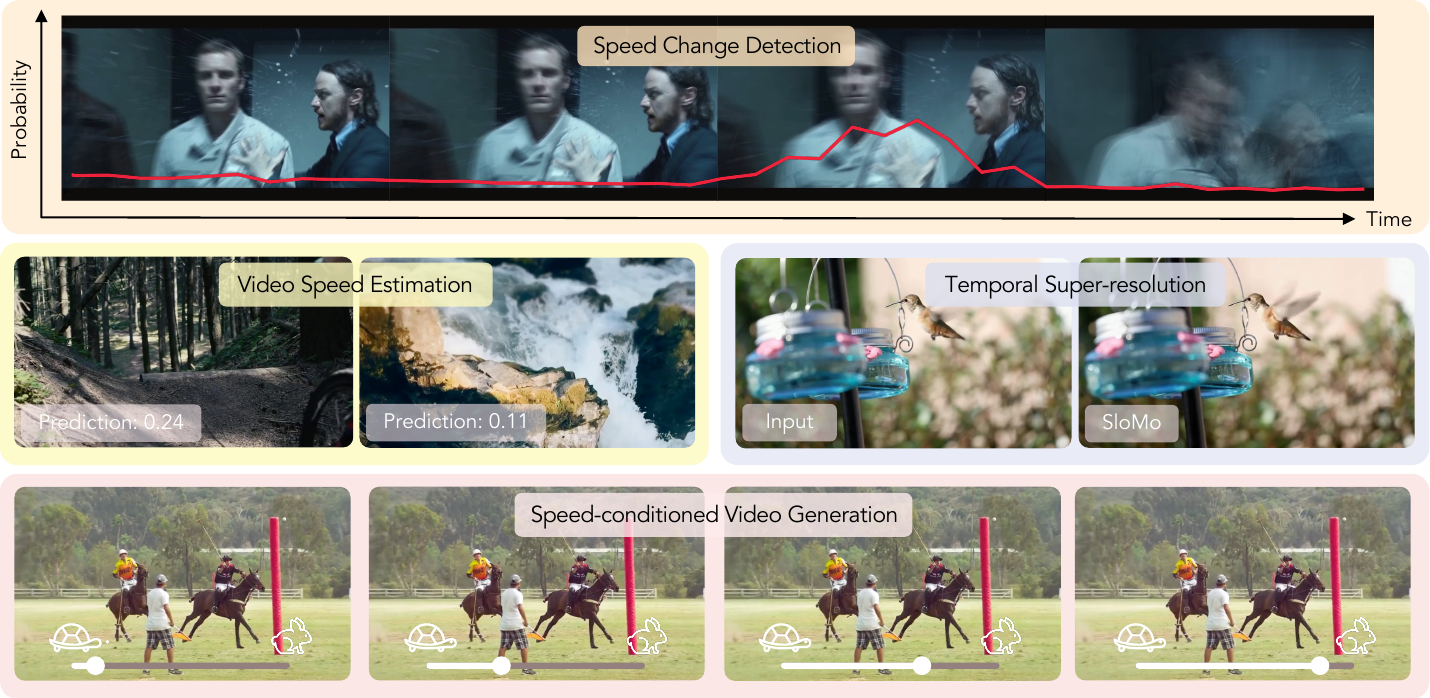}
    }
    \vspace{-5mm}
    \caption{
        We develop models towards \textbf{perceiving and manipulating the flow of time}, including
        (a)~\emph{speed-change detection}, which locates moments where playback speed shifts; 
        (b)~\emph{video speed estimation}, which infers how much a video has been sped up or slowed down; 
        (c)~\emph{extreme temporal super-resolution}, converting low-FPS, blurry videos into their high-FPS, clear counterparts; and 
        (d)~\emph{speed-conditioned video generation}, synthesizing the same event at user-specified temporal speeds. For more details on these results, please refer to \cref{sec:experiments} and the project page.
        \emph{Please open in Adobe Acrobat Reader to view the embedded animations.}
        }
    \label{fig:teaser}
\end{figure}

\vspace{-6mm}
\section{Introduction}
\vspace{-1mm}
\label{sec:intro}

Humans exhibit a strong intuition for the flow of time. Even without explicit cues, we often sense when a video is sped up or slowed down: surface ripples propagating too quickly, objects falling at unnatural rates, or human motion that feels sluggish. This intuition reflects an implicit understanding of the speed at which events unfold in the physical world~\cite{steinhof2025time}. Despite remarkable progress in video understanding and generation, current models lack such temporal reasoning~\cite{li2023videochat, song2024moviechat, lin2024video}. When asked to predict playback speed, modern vision–language models (VLMs)~\cite{comanici2025gemini, zhu2025internvl3}  frequently hallucinate or give incorrect predictions. When prompted to generate content at a particular speed, existing generative models~\cite{wan2.1, wan2025wan, ldmvfi} often ignore instructions or generate video with imprecise timing. These issues highlight a fundamental limitation in current models: the inability to reason about time.

This limitation is not particularly surprising. Modern models primarily learn from videos captured at standard frame rates (\emph{e.g.}, 24–60 fps)~\cite{chen2024panda, wang2023internvid}, and therefore never learn the notion of temporal speed.  In effect, we are asking the model to infer time’s variability from observing a single cadence. We argue that for models to understand the flow of time and to control temporal dynamics, they must be exposed to data spanning the continuum of temporal speeds.

Obtaining such data at scale is not trivial. 
Although there are vast untapped stores of slow-motion videos online, the associated metadata required to train models, such as the frame rate, playback speed, and time stamps, are often incomplete. A natural solution would be to have humans manually annotate the videos.  
However, naively labeling is prohibitively time consuming and imprecise. Therefore, we look to other directions for providing temporal supervision for video understanding and generation models.

\begin{figure}[t]
    \centering
    \includegraphics[width=\linewidth]{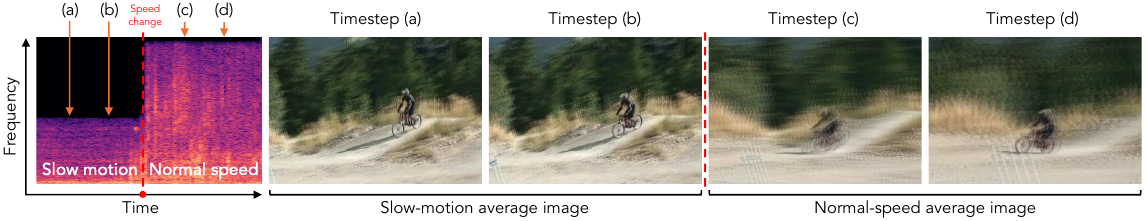}
    \vspace{-6mm}
    \caption{\textbf{Audio signal naturally encodes cues for speed change detection.} 
    When a video's playback speed changes, its audio pitch shifts accordingly -- speeding up raises the pitch, while slowing down lowers it. This provides \textbf{free cross-modal supervision} for detecting speed changes by locating moments of pitch variation. Left: Audio spectrogram ({\textbf{used only during training}}).
    Right: Average image at each selected timestep, obtained by averaging nearby frames. Higher playback speed leads to larger content changes and therefore greater blur.
    }
    \label{fig:speed-change}
\end{figure}

In this paper, we show that temporal cues and multi-modal signals in videos can be leveraged to infer playback speed without explicit labels.
Our first key insight builds on the principle of \emph{time–frequency scaling}: when a video's playback speed changes, its audio pitch shifts. 
Thus, by locating regions where the pitch shifts (Fig.~\ref{fig:speed-change}), we can obtain free cross-modal supervision to train models that visually detect speed changes. 

Our second insight exploits the \emph{equivariance} of speed estimation under temporal rescaling: if we downsample a video by a certain factor, its perceived speed should decrease proportionally by the same factor. By resampling each clip and enforcing this proportional relationship in the model’s predictions, we use temporal resampling as a powerful self-supervised training signal. 
Together, these complementary cues enable models to infer the lapse of time directly from visual dynamics, even in the absence of clean or reliable temporal labels.

We apply our temporal speed models to in-the-wild videos collected from online platforms.
Using these models, we automatically annotate and curate a large-scale slow-motion dataset spanning diverse activities, motion patterns, and temporal scales.
These slow-motion videos, typically captured at high frame rates (\eg, 1000+ FPS) and played back at standard frame rates (\eg, 30 FPS), preserve fine temporal details and exhibit reduced motion blur, allowing machines to perceive and learn how real-world dynamics actually unfold.

Next, using our newly constructed dataset, we develop models capable of manipulating the flow of time in videos.
We focus on two tasks: (i) speed-conditioned video generation, which equips models with the ability to render physical motion at different temporal speeds; and (ii) temporal super-resolution, which converts videos into higher frame-rate sequences. 
Both tasks are challenging: standard videos often lack fine-grained motion cues due to motion blur, and synthesizing motion at precise speeds requires accurate temporal modeling.
We show that finetuning existing models on our slow-motion dataset enables them to generate lifelike dynamics at controllable playback speeds and to interpolate frames at higher effective frame rates.

Since the tasks explored in this work are relatively new, there are no established benchmarks or standard evaluation protocols. To ensure a fair and meaningful assessment, we carefully curate dedicated evaluation datasets and conduct comprehensive human perceptual studies.
Our experiments show that the proposed models perform strongly across both understanding and generation tasks, achieving 92\% accuracy on speed-change detection, near-human accuracy on playback-speed estimation, 80.3\% human-preference win rate over the base model for temporal super-resolution.
\section{Related Works}

\myparagraph{Time as a learnable visual concept.}
Humans exhibit a remarkably malleable perception of time when watching videos. Studies in cognitive science~\cite{steinhof2025time} show that people tend to overestimate the duration of an action when it is presented in slow motion. From a computer vision perspective, learning the concept of time has long intrigued researchers.
For instance, earlier works on arrow-of-time (AoT)~\cite{seeing-arrow-of-time,learning-arrow-of-time,ghodrati2018video} sought to determine if a video is playing forward or backward. Building on this line of research, recent studies have extended large multimodal models with AoT awareness~\cite{xue2025seeing,azzolini2025cosmos} as well as broader temporal reasoning capabilities~\cite{bagad2023test,wang2023paxion,du2024reversed,ding2025language}.

In the context of video representation learning, temporal information has also been widely exploited as a self-supervised signal~\cite{misra2016shuffle,dorkenwald2022scvrl,lee2017unsupervised, benaim2020speednet,yao2020videoplaybackperception,wang2020self}. In particular, pace classification has emerged as an effective pretext task for improving temporal understanding, where videos are augmented with different temporal sampling rates and models are trained to classify them as slow, normal, or fast ~\cite{benaim2020speednet,yao2020videoplaybackperception,wang2020self}.
Extending this idea, SpeedNet~\cite{benaim2020speednet} demonstrates that the learned pace classifier can further be leveraged to estimate “speediness” scores across a video, enabling adaptive video speedups.

More recently, Pulse-of-Motion~\cite{gao2026pulse} proposed a video FPS predictor to evaluate the speed correctness of major video generation models.
However, their speed-perception model is trained primarily on standard frame-rate videos ($\le 60$ fps), with only 8 videos at 120 fps (UVG~\cite{mercat2020uvg}) and 118 videos at 240 fps (Adobe240fps~\cite{adobe240}). As a result, their notion of speed is limited to a narrow range (1/2, 1, 2, 4$\times$) and oftentimes inaccurate, which we will show in Sec. \ref{sec:exp-playback-speed-estimation}.
In contrast, our method is trained on videos spanning a wide range of speeds and uses a dedicated self-supervised objective, allowing it to predict speed across a wide range of temporal scales.

\myparagraph{Manipulating the continuum of time.} 
Understanding and altering the temporal dimension of videos is a fundamental problem in computer vision.
Frame interpolation addresses this by synthesizing intermediate frames between two existing inputs, effectively increasing a video's temporal resolution.
Earlier approaches to interpolation primarily relied on kernels~\cite{liu2017video,niklaus2017video,xiang2020zooming} or optical flow estimation~\cite{huang2022rife,film} to model motion, whereas recent generative approaches leverage the rich priors of image or video diffusion models to better handle complex motion and improve realism~\cite{ldmvfi, jain2024video,generative-inbetween, feng2024explorative}.

Beyond interpolation, prior work on video generation has explored motion control techniques~\cite{blattmann2023stable, wang2024motionctrl, burgert2025go, ati} to explicitly manipulate temporal dynamics. More recently, video editing approaches like BulletTime~\cite{wang2025bullettime} and SpaceTimePilot~\cite{huang2025spacetimepilot} have enabled joint editing of camera motion and temporal progression, offering flexible spatio-temporal control.

In this work, we focus on playback speed-conditioned video generation, ranging from normal speed (1.0$\times$) to extreme slow motion (0.01$\times$). Unlike editing-centric settings~\cite{wang2025bullettime,huang2025spacetimepilot} that merely rescale the temporal axis of an existing video, our generation-centric task requires the model to internalize the real-world progression speed of physical motion and accurately fabricate it at the requested speed (see supp.~for detailed comparisons). Furthermore, the broad conditioning range requires our model to generate dynamics at an exceptionally high temporal resolution. By leveraging our proposed \dataset, which spans diverse temporal scales and real-world motion patterns, our model achieves high-fidelity video generation across a wide range of speeds, outperforming existing approaches trained only on low-frame-rate or synthetic data (Sec. \ref{sec:experiments-speed-controlled}).

\myparagraph{Slow-motion video datasets.}
Large-scale video generation datasets such as WebVid-10M~\cite{webvid-10m}, Panda-70M~\cite{chen2024panda}, and OpenVid-1M~\cite{openvid} have driven recent advances in video generation. However, these datasets primarily contain standard-speed footage (24–60~fps) and thus lack the dense temporal continuity required to model realistic motion dynamics. In contrast, slow-motion videos, recorded at hundreds or even thousands of fps, offer much finer temporal details that normal-speed videos cannot capture. Existing slow-motion datasets, however, remain limited in either scale, frame rate, or scene diversity. For example, Adobe240fps~\cite{adobe240}, YouTube240~\cite{youtube240}, and NfS~\cite{nfs} each contain only a few hundred videos at 240~fps; SportsSloMo~\cite{chen2023sportsslomo} includes more clips but focuses solely on sports scenes; and X4K1000FPS~\cite{sim2021xvfi} consists of 1000~fps self-recorded videos yet remains small in size. Consequently, these datasets primarily serve as benchmarks for video frame interpolation rather than as large-scale resources for model training. In contrast, our proposed~\dataset~is the first large-scale and general-purpose slow-motion dataset, offering a foundation for learning fine-grained dynamics.

\begin{table}[t]
  \centering \small
    \scalebox{1.0}{
	\begin{tabular}{l|ccccccc}
		\toprule
		Dataset & ~~\#Clips~~ & ~\#Videos~ & ~\#Frames~ & ~Max raw FPS~ & ~Content~  \\
		\midrule
		Adobe240fps~\cite{adobe240} & 118 		& 118	  & 80K   & 240     &  Urban \\
		YouTube240~\cite{youtube240} 	& 1{,}014 	& - & 296K  & 240     &  Unknown \\
		NfS~\cite{nfs} 	    & 100 		& 100  	  & 383K  & 240     & Generic \\
		X4K1000FPS~\cite{sim2021xvfi} 	& 175$^{*}$     	& 175     & 875K  & 1{,}000   & Urban \\
		SportsSloMo~\cite{chen2023sportsslomo} & 8,498$^{*}$     & 259     & 1.2M  & 240	    & Sports \\
		\midrule
		\dataset (Ours)~~	& 44{,}632	& 18{,}235  & 18M & 10{,}000+ 	& 	 Generic \\
		\bottomrule
	\end{tabular}
    }
    \caption{\textbf{Comparison with existing high-frame-rate datasets.}
    Our dataset is the largest generic slow-motion collection to date, containing over $70\times$ more videos and $150\times$ more frames than previous datasets. 
    $^{*}$Counts are based on full-length clips rather than short, fixed-duration trimmed segments.}    
  \label{tab:datasets}
\end{table}

\section{Perceiving the Flow of Time}
\label{sec:understanding}
We aim to develop computational visual systems that can perceive and manipulate the flow of time. 
In this section, we focus on how models understand and perceive time, answering fundamental questions such as \emph{Has this video been sped up or slowed down? If so, when, and by how much?}
These abilities form the foundation for deeper temporal reasoning and pave the way for fine-grained temporal control, which we explore in Sec. \ref{sec:manipulation}.

As most of the metadata associated with public slow-motion videos, such as playback speed, is often noisy or incomplete, data for models to learn from is lacking.   
In the following section, we detail how models can learn to detect speed changes and infer the speed of time through self-supervision.
These models then curate the largest slow-motion video dataset to date for visually understanding and controlling the flow of time.

\subsection{Learning to Detect Temporal Speed Changes}
\label{sec:speed_change}
We begin with detecting temporal speed changes in videos.  
Our goal is to locate times in the video where the playback speed changes, transitioning between normal, fast, and slow motion.
Such effects are common in modern videos and are often used to emphasize key moments or create dramatic effects.
If a model can reliably detect these changes, it not only reveals an understanding of temporal continuity, but also provides a foundation for estimating absolute speed and calibrating temporal labels in large-scale video datasets.

Manually annotating speed changes is laborious and difficult to scale, and using optical flow as a proxy measurement is often noisy and unreliable (see Sec.~\ref{sec:experiments}). We instead leverage cues encoded in the audio of videos. Specifically, our key idea is to use the principle of \textbf{time–frequency scaling} to detect speed changes. For videos that contain original audio tracks (\emph{e.g.}, not dubbed or overlaid with background music), there exists a natural coupling between visual motion and sound.
When such videos are sped up (compressed in time), their corresponding audio frequencies shift upward, resulting in a higher pitch, and vice versa when slowed down. 
These salient audio cues (pitch shift) allows us to effectively locate speed-change events. An illustration is shown in Fig. \ref{fig:speed-change}.

Using this strategy, we automatically collect over 8K speed-change labels. We then finetune a \emph{visual} speed-change detector based on VideoMAEv2~\cite{wang2023videomae} using a standard binary cross-entropy loss. Notably, during inference, the detector \emph{operates solely on visual input} and is agnostic to \emph{audio} content.

\subsection{Learning to Infer the Speed of Time}

\begin{wrapfigure}{r}{0.485\textwidth}
    \centering
    \includegraphics[width=.98\linewidth, trim=0.4cm 0cm 0.5cm 0.4cm]{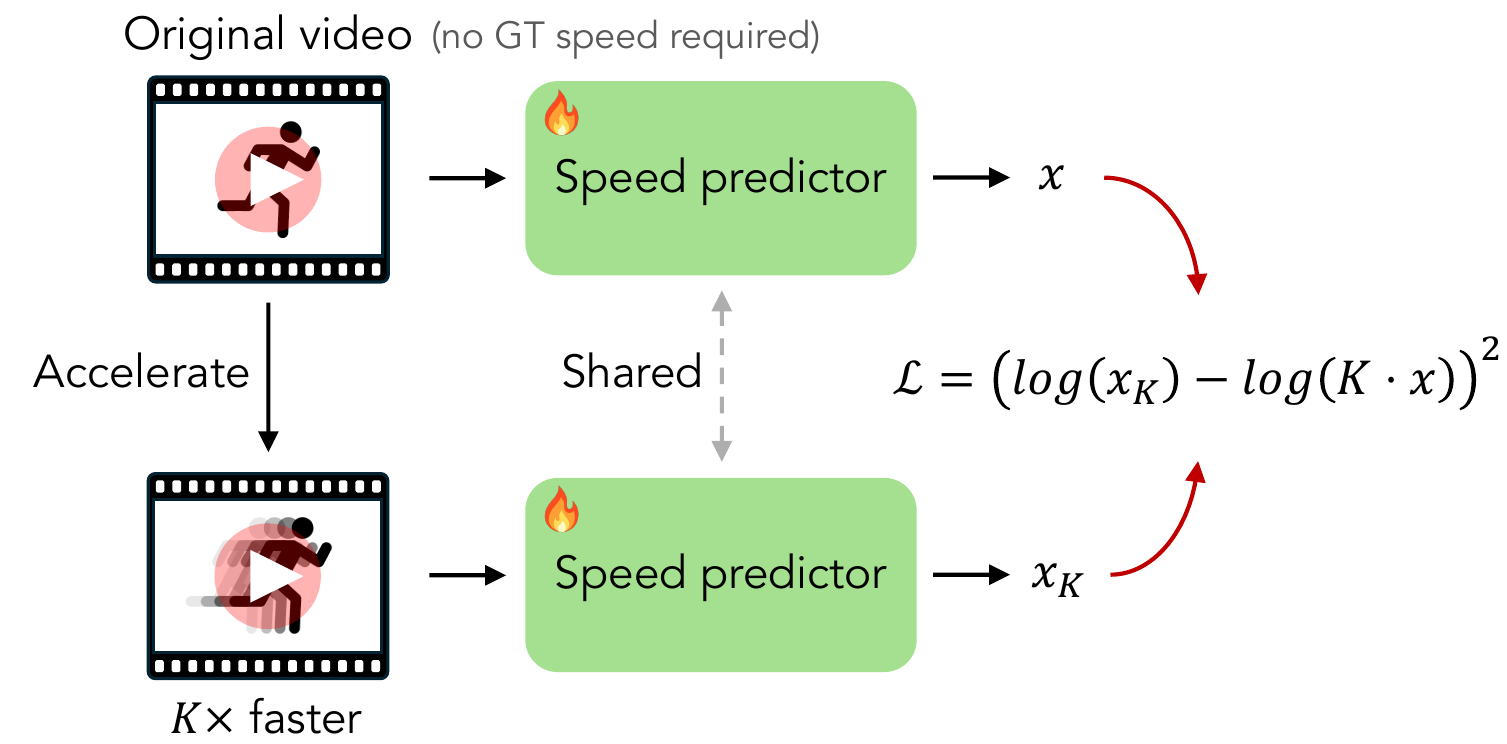}
    \caption{
    \textbf{Learning to predict speed.}
    Our speed estimator is trained with both self-supervised and supervised objectives.  
    For videos without ground-truth speed, we enforce temporal consistency by subsampling the video by a factor of $K$, feeding both original and accelerated clips into the model, and constraining their predicted speeds to differ by $K$.
    For videos with known frame rates, we directly regress the playback speed.
    }
    \label{fig:self-supervised-learning}
\end{wrapfigure}

Once we can detect speed changes within videos, a natural follow-up question arises: By how much has the video been sped up or slowed down?

As there is no current reliable source of supervision, we rely on a self-supervised objective. Our key insight is that the speed estimation model should be \textbf{equivariant} to temporal resampling, \emph{i.e.,} if we speed up a video by a factor of $k$, the predicted speed should scale by $k$. By enforcing this proportional relationship between the model's input and output, we can transform temporal resampling into a powerful self-supervised training signal (\cref{fig:self-supervised-learning}). Formally, let $\mathbf{V}$ be a video clip sampled from a video of duration $T$, and $\mathbf{V}^k$ be the $k$-times accelerated version of $\mathbf{V}$, where $k \sim \mathcal{N}(1, \tfrac{T}{2})$. Denote $f_{\theta}$ as the playback speed estimator. We train the model $f_{\theta}$ using:
{
\setlength{\abovedisplayskip}{6pt}
\setlength{\belowdisplayskip}{6pt}
\begin{equation}
{\mathcal L}
=\! \big[\log f_{\theta}(\mathbf{V}^k)-\log\!\left(k \cdot f_{\theta}(\mathbf{V})\right)\big]^2.
\end{equation}
}

\myparagraph{Calibration.} While the self-supervised loss allows the model to predict playback speed purely from motion cues without relying on ground-truth labels, the estimation is sometimes only correct up to a scale. To anchor predictions to absolute playback rates, we additionally incorporate a small set of videos with known ground-truth playback speeds. For example, we use videos from the Adobe240FPS dataset~\cite{adobe240}, where the playback speed is precisely determined. In addition to the self-supervised loss, we train the network with small amounts of supervised data in log space.

\begin{table}[t]
    \centering
        \centering\small
        \begin{tabular}{l|cccc}
        \toprule
        Method & ~$\rho\uparrow$~ & $r_s\uparrow$ ~& RMSE$\downarrow$~ & $e^{\text{RMSE}}\downarrow$ \\
        \midrule
        Human expert      & \textbf{0.880}    & \textbf{0.783}   & \textbf{0.492}   & \textbf{1.636}   \\
        \midrule
        Optical flow & 0.385        & 0.354        & -		 	 & -		  \\
        VideoLLM~\cite{comanici2025gemini}         & 0.426        & 0.308        & 1.568      & 4.796      \\
        SpeedNet~\cite{benaim2020speednet}        & 0.476        & 0.331        & 1.261 & 3.529     \\
        Pulse-of-Motion$^{*}$~\cite{gao2026pulse}        & 0.508        & 0.525        & 1.181 & 3.258     \\ 
        \textbf{Ours}    & \textbf{0.735}   & \textbf{0.706}   & \textbf{0.649} & \textbf{1.913} \\ 
        \bottomrule
        \end{tabular}
        \caption{{\bf Video speed prediction results.}
        We report the Pearson and Spearman correlation coefficients ($\rho$, $r_s$), root mean squared error (RMSE), and $e^{\mathrm{RMSE}}$, all computed in log space.
        Because optical-flow magnitude is used only to rank relative speed, its RMSE is omitted.
        Our method significantly narrows the gap between machine and human performance.
        $^{*}$Results are obtained using the authors' publicly available checkpoint.
        }        
        \label{tab:speed_prediction}
\end{table}

\myparagraph{Iterative prediction.} With calibration, our model can already produce relatively accurate speed estimations.
However, in practice, we find that our model tends to underestimate when encountering extremely slow-motion videos. We hypothesize that this is because the motion difference in ultra-slow videos are often very subtle due to resolution constraints. 
Subsampling frames before inference further makes it challenging.
To mitigate the issue, we adopt an iterative prediction approach. 
If a video is initially predicted to have a playback speed of $x$, we accelerate the video accordingly to bring it closer to normal speed, and then re-estimate its playback speed to obtain a more precise prediction. Repeating this process iteratively produces progressively refined predictions. 
The intuition is that speed differences become more discernible when the playback speed approaches one, where more training data is also concentrated, allowing the model to estimate speed more reliably.
We empirically unroll the estimator three times. 
Please refer to the supp.~material for more details and ablation study.

\subsection{Annotating the Speed of Time}
\label{sec:dataset}
With reliable speed understanding models in hand, we now leverage them to transform in-the-wild slow-motion videos into a high-quality annotated dataset.

\myparagraph{Sources.} 
We source videos from YouTube, Vimeo, and Flickr using queries such as ``high frame rate,'' ``high-speed camera,'' and ``slow motion,'' along with their synonyms. 
Following \cite{blattmann2023stable,luo2025beyond}, we then use TransNetv2~\cite{soucek2024transnet} to segment videos into shots and an OCR model~\cite{ye2023dptext} to remove video clips with excessive text overlay. 
To further improve quality, we filter out CGI or screen recordings using Qwen2.5-VL~\cite{bai2025qwen2} and discard low-quality samples based on video quality assessment~\cite{he2024cover}.

\myparagraph{Speed annotation.} 
As many videos have unspecified or heterogeneous playback speed, we leverage our speed change detector to segment videos into clips with homogeneous playback speeds. We then run our speed estimator to annotate the playback speed for each clip. Finally, we densely caption them with InternVL3~\cite{zhu2025internvl3}. We refer the readers to the supp.~material for more details.

\myparagraph{\dataset.} Our resulting dataset contains 44{,}632 slow-motion video clips with a total of 18 million frames, making it the largest of its kind. 
It spans diverse real-world scenes and motion patterns, and covers a broad range of video durations.  A comparison with previous datasets is shown in Tab. \ref{tab:datasets}. For more statistics, please see the supp.~material.

\section{Manipulating the Flow of Time}
\label{sec:manipulation}
Having learned to detect speed changes and infer the speed of time in videos, we now turn to the problem of manipulating it.  
In this section, we demonstrate how models can learn to adjust the dynamics of generated content or to slow down existing videos with fine-grained precision.

\begin{figure*}[t]
    \centering
    \includegraphics[width=\linewidth]{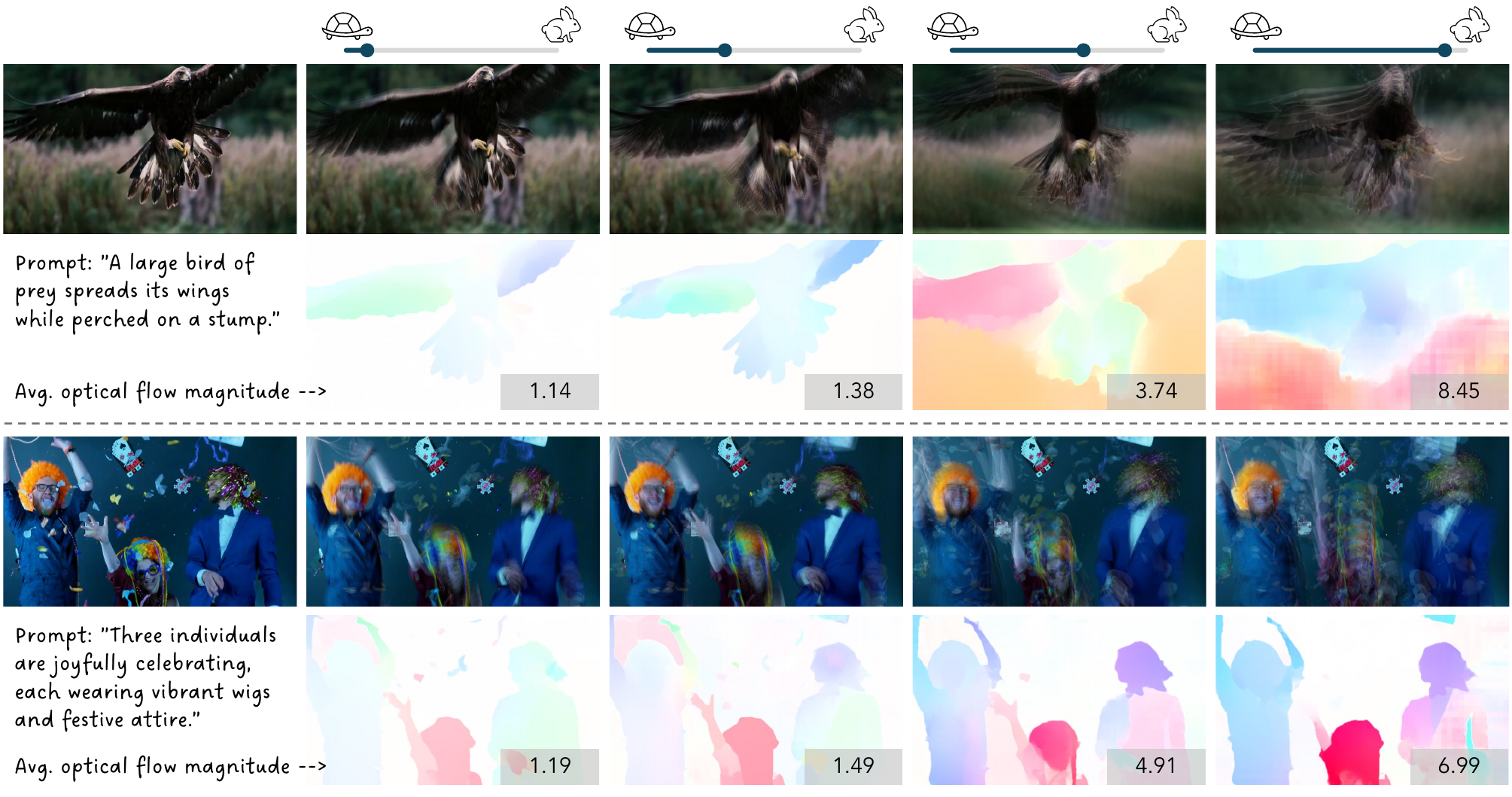}
    \vspace{-5mm}
    \caption{\textbf{Speed-conditioned video generation.} Given a text prompt and an image as input (first column), our model learns to generate videos of different speed.
    The first row shows the average image of the video, while the second row visualizes the optical flow between the first and second frame. 
    When the speed control is set to slow, the average image should still be crisp (because the scene does not change much) and the magnitude of optical flow is small (and thus the color is faint). As the speed of the video increases, the average image becomes blurrier and the magnitude of optical flow increases (the color become more saturated). For more results, see the project page.}
    \label{fig:i2v-vis}
\end{figure*}

\subsection{Speed-conditioned Video Generation}
Given an image, a text prompt, and a target playback speed, our goal is to develop a model that can generate dynamic visual content unfolding at the desired temporal rate and show physical dynamics (\eg, fluid dynamics, shattering, vibration, etc) perceptible only at that temporal scale.

Intuitively, for any text-to-video model, a potential way to control playback speed would be to incorporate motion-related text modifiers (\emph{e.g.}, \texttt{ultra-slow}, \texttt{slightly-slow}) into the prompt. However, as we show in Sec.~\ref{sec:experiments-speed-controlled}, textual conditioning alone is insufficient for generating videos at the desired speed, often failing to make any noticeable difference. We conjecture that this is because physical motion cues are entangled with other semantic factors in the text prompt, resulting in a weak and easily overloaded conditioning signal. 

With this in mind, we propose to leverage our speed-annotated dataset to train a video model with \emph{explicit speed control}. 
We build model upon Wan2.1-I2V~\cite{wan2.1} with two key modifications. 
Given a target speed, we first discretize it into logarithmically spaced buckets that represent different temporal speeds. Empirically, these buckets span from extremely slow (0.01$\times$) to normal speed (1.0$\times$), covering the range in which most \dataset videos fall (see supp. for analysis). Then we encode the bucket id with sinusoidal positional embedding $\phi$, apply a multilayer perceptron $\mathrm{MLP}_{\theta}$, and add it to the timestep embedding \cite{bahmani2025ac3d}:
{
\small
\setlength{\abovedisplayskip}{6pt}
\setlength{\belowdisplayskip}{-1pt}
\begin{align}
\texttt{Bucket\_ID} =\left\lfloor
\frac{\log(\texttt{speed}) - \log(0.01)}{\log(1) - \log(0.01)} \cdot N_{\text{buckets}}\right\rfloor, \nonumber\\
\texttt{time\_emb} \leftarrow \texttt{time\_emb} + \mathrm{MLP}_{\theta}\cdot\phi(\texttt{Bucket\_ID}).
\end{align}
}

This encourages the model to align its denoising schedule with the temporal speed of the video. Empirically, we set $N_{\text{buckets}}=10$.  

To further enhance the control over temporal speed, we modulate the latent features with playback speed through frame-wise conditioning \cite{bahmani2025ac3d}: 
{
\small
\setlength{\abovedisplayskip}{6pt}
\setlength{\belowdisplayskip}{-1pt}
\begin{align}
\texttt{latent}[i] &\leftarrow \texttt{latent}[i] + \mathrm{MLP}_{\psi}(\phi(i \cdot \texttt{speed})),
\end{align}
}

where $\texttt{latent}[i]$ denotes the latent feature at temporal index $i$. 

We follow the image-to-video setup of Wan2.1-I2V~\cite{wan2.1}.
During training, we optimize both the linear projection layers and the LoRA adapters applied to the transformer backbone. The resulting model supports controllable slow-motion synthesis under text, image, and speed conditions.

\subsection{Extreme Temporal Super-Resolution}
\label{sec:temporal-super-res}
Finally, we apply our data to the task of temporal super-resolution, where the goal is to transform low-FPS, blurry videos into their high-FPS, clear counterparts. While previous approaches typically assume clean and sharp input frames~\cite{ldmvfi,film,generative-inbetween}, we focus on a more practical and challenging setting in which the input may have motion blur. Low-FPS videos naturally exhibit stronger motion blur due to longer exposure times, making the reconstruction of high velocity details significantly harder. 

Our key idea is to use our large-scale slow-motion dataset to synthesize a wide range of low-FPS, motion blurred videos for training. 
Following~\cite{brooks2019learning,adobe240,nah2017deep}, we generate synthetic blurry inputs by averaging a temporally centered window of 8 frames, followed by temporal subsampling to mimic low-frame-rate capture. 

We build our model upon Wan2.1-VACE~\cite{jiang2025vace} for its flexible conditioning capability. The VACE framework takes a reference video and an arbitrary binary mask that specifies the frames or regions to be generated. 
We fine-tune LoRA adapters on top of the pretrained base model.
By training on these paired data, the model can jointly perform motion deblurring and frame interpolation, faithfully capturing real-world shooting conditions. Notably, it also achieves superior performance even in the traditional setting with clean, non-blurry inputs. Due to computational resource constraints, we focus on 8-times upsampling.

\section{Experiments}
\label{sec:experiments}
In this section, we evaluate how well our proposed techniques perceive and manipulate the flow of time. 
We first benchmark our speed-change detector and playback-speed estimator, and then showcase that models trained on our dataset exhibit improved speed controllability and temporal super-resolution capabilities.

\begin{figure}[t]
    \centering
    \includegraphics[width=\linewidth]{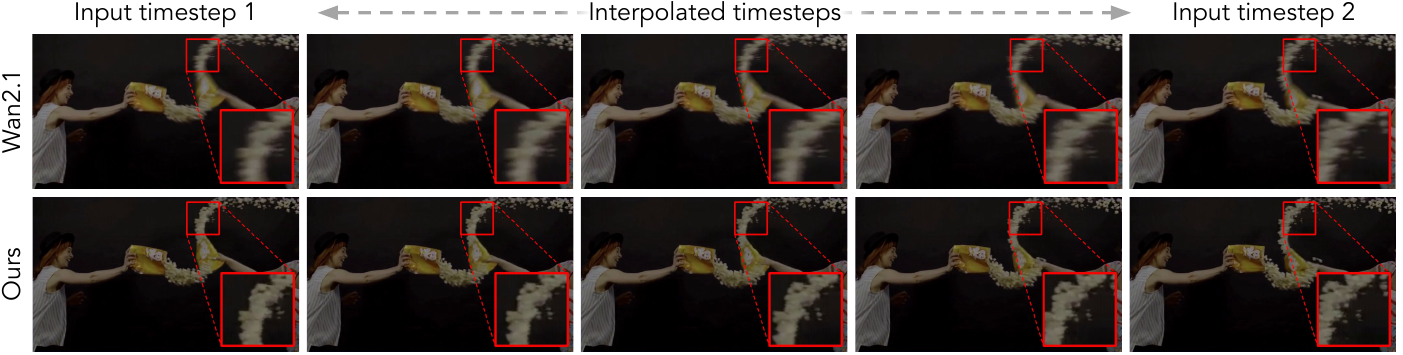}
    \vspace{-5mm}
    \caption{\textbf{Temporal super-resolution qualitative results.}
    We compare against the strongest baseline, Wan2.1, under the \emph{Blurred-input} setting. Given blurry input frames, the baseline produces blurry intermediate frames, whereas our method generates sharp frames with smooth motion, closely resembling footage captured by a high-speed camera.
    Inset zooms are provided for clearer visual comparison.}
    \label{fig:vif-vis}
\end{figure}

\subsection{Speed Change Detection}

\myparagraph{Data.} 
We evaluate our model on the test split of our curated speed-change data.
Because these labels are automatically derived from audio cues and may contain false positives, we have four annotators re-label them and keep only samples where human annotations and audio-based labels agree. We sample 2-second clips from test videos. 
If the speed change occurs between $\frac{1}{3}$ and $\frac{2}{3}$ of the clip, we treat it as a positive sample; if the speed change occurs elsewhere, or if no speed change is present, we treat it as a negative sample.

\myparagraph{Baselines.}
We compare against a SOTA VideoLLM, Gemini 2.5~\cite{comanici2025gemini}, as well as a flow-based detector. For the latter, we adopt SEA-RAFT~\cite{sea-raft} to compute flow magnitudes between adjacent frames, smooth the magnitudes with a moving window of size 5, and classify a clip as positive if the magnitude within its middle third changes by more than a threshold determined based on a held-out set.

\myparagraph{Results.} Our model achieves the highest test accuracy ($92.4$\%), outperforming both Gemini 2.5 ($59.5$\%) and the flow-based baseline ($80.4$\%).
We further apply our model to an iconic movie scene in \textit{X-Men} with time-freeze effects (first row of \cref{fig:teaser}). We visualize model predictions alongside movie frames at corresponding timesteps.
To show motion intensity at each moment, we blend each frame with its neighbors.
Here, time accelerates from slow motion to normal speed, causing initially sharp frames to become blurry.
Our model accurately identifies this transition moment, demonstrating its potential for video forensics.


\subsection{Playback Speed Estimation}
\label{sec:exp-playback-speed-estimation}

\begin{table}[t]
    \centering\small
    \begin{tabular}{l|cccccccc}
        \toprule
        \multirow{2.5}{*}{Method} & \multicolumn{4}{c}{{\bf DAVIS}} & \multicolumn{4}{c}{{\bf ~\dataset-Test}}\\
        \cmidrule(r){2-5}\cmidrule(r){6-9}
        & ~FloLPIPS$\downarrow$~ & ~LPIPS$\downarrow$~ & ~FID$\downarrow$~ & ~FVD$\downarrow$~ & ~FloLPIPS$\downarrow$~ & ~LPIPS$\downarrow$~ & ~FID$\downarrow$~ & ~FVD$\downarrow$~  \\
        \midrule
        FILM~\cite{film} & \underline{0.252} & \bf{0.200} & \underline{20.7} & 711.3 & \underline{0.087}     & \underline{0.066} & \bf{10.7} & \underline{257.6} \\
        LDMVFI~\cite{ldmvfi}  & 0.307 & 0.251         & 31.6 & 916.0 & 0.139     & 0.113 & 22.7 & 443.0  \\
        GI~\cite{generative-inbetween}      & 0.353         & 0.234 & 21.4 & \underline{552.1} &  0.124     & 0.093 & 14.9 & 503.6 \\
        Wan2.1~\cite{jiang2025vace}
        & 0.316         & 0.260 & 24.7 & 634.5 & 0.108     & 0.078     & 16.6 & 594.0 \\    
        \midrule
        Ours & \bf{0.242} & \underline{0.203} & \bf{18.2} & \bf{394.0} & \bf{0.078} & \bf{0.061} & \underline{10.9} & \bf{182.2} \\    
        \bottomrule
    \end{tabular}
    \caption{\textbf{Temporal super-resolution from clear inputs.} 
    Following prior work on frame interpolation~\cite{film,ldmvfi,generative-inbetween}, we set aside the motion blur typical of low-FPS capture and evaluate under \emph{clean} inputs.  
    Specifically, we obtain low-FPS inputs by $8\times$ temporal subsampling and task models with reconstructing the original videos. 
    Finetuned on \dataset, our model outperforms prior methods on \emph{video-based} metrics (\ie, FloLPIPS, FVD) and matches or achieves the best results on \emph{image-based} metrics (\ie, LPIPS, FID). 
    Best results are in \textbf{bold} and second-best are \underline{underlined}.
    }
    \label{tab:results-woblur}
\end{table}
    
\begin{table}[t]
    \centering\small
    \begin{tabular}{l|cccc}
        \toprule
        Method & FloLPIPS$\downarrow$ & LPIPS$\downarrow$ & FID$\downarrow$ & FVD$\downarrow$ \\
        \midrule
        FILM~\cite{film} & 0.099 & 0.080 & 20.3 & 250.0  \\
        LDMVFI~\cite{ldmvfi} & 0.136 & 0.117 & 31.4 & 340.4   \\
        GI~\cite{generative-inbetween} & 0.123 & 0.108 & 26.1 & 439.4  \\
        Wan2.1~\cite{jiang2025vace}
        & 0.126 & 0.095      & 28.5 & 436.1  \\    
        \midrule
        \bf{Ours} & \bf{0.067} & \bf{0.058} & \bf{12.4} & \bf{134.3} \\    
        \bottomrule
    \end{tabular}   
    \caption{\textbf{Temporal super-resolution results on motion-blurred videos (SloMo-44K-Test).} 
    Our method achieves the best performance across all metrics, demonstrating robust temporal super-resolution under motion blur, a common artifact in real-world low-FPS videos.
    }    
    \label{tab:results-blur}
\end{table}

\begin{figure}[t]
    \centering
    \includegraphics[width=\linewidth]{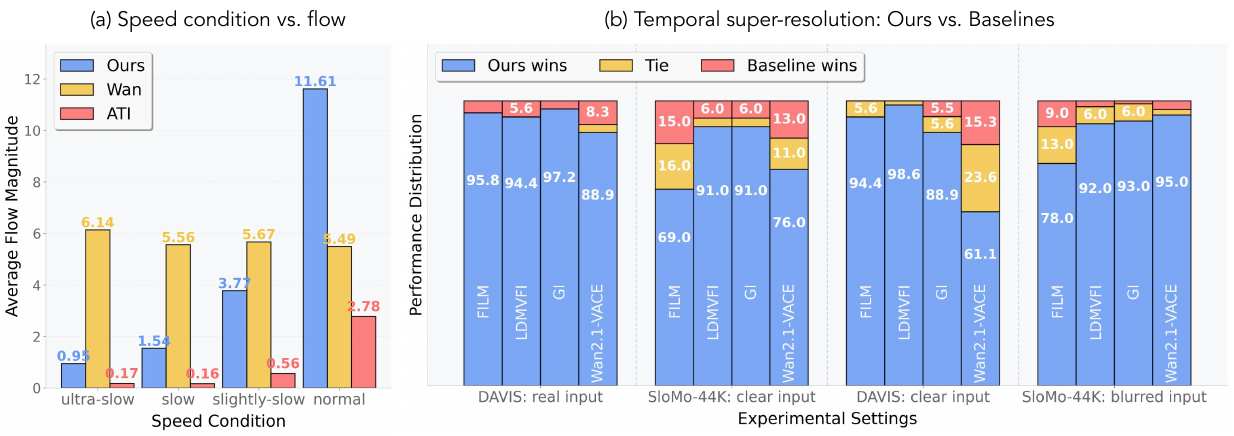}
    \vspace{-6mm}
    \caption{
    \textbf{(a) Speed condition vs. optical flow.}
    Our speed control strongly correlates with the average optical flow magnitude of videos generated under different speed conditions, whereas the baselines yield similar motion magnitudes for slow and ultra-slow speeds.
    \textbf{(b) User study on temporal super-resolution.}
    The results show that human users consistently prefer videos generated by our model (blue) over those generated by baselines (red) across all settings.
    }
    \label{fig:results}
\end{figure}

\begin{table}[t]
    \centering
    \begin{minipage}[t]{0.48\linewidth}
        \vspace{0pt}
        \centering\small
        \begin{tabular}{l|cc}
        \toprule
        Method & ~FID$\downarrow$~ & ~FVD$\downarrow$~ \\ \midrule
        ATI~\cite{ati} & 73.4 &	1473.5  \\
        Wan2.1~\cite{wan2.1} & 72.2 &	1266.8  \\
        \textbf{Ours} & \textbf{68.4} &	\textbf{1114.1} \\ \bottomrule
        \end{tabular}
        \caption{\textbf{Speed-controlled video generation results.} We compare our video model trained on \dataset with the pretrained Wan2.1~\cite{wan2.1} and the trajectory-based model, ATI~\cite{ati}. Our model demonstrates superior quality in slow-motion generation.}    
        \label{tab:i2v-results}
    \end{minipage}
    \hfill
    \begin{minipage}[t]{0.48\linewidth}
        \vspace{0pt}
        \centering\small
        \begin{tabular}{l|cc}
        \toprule
        Training data & ~FID$\downarrow$~ & ~FVD$\downarrow$~ \\ \midrule
        Standard videos & 72.4 &	1392.9  \\
        \textbf{\dataset} & \textbf{68.4} &	\textbf{1114.1} \\ \bottomrule
        \end{tabular}
        \caption{\textbf{Speed-conditioned video generation with different training data.} We train our video generation model on either \dataset or standard-FPS datasets. Results show that training on \dataset yields superior slow-motion generation quality, confirming the importance of high-frame-rate data.}
        \label{tab:i2v-dataset}
    \end{minipage}
\end{table}

\myparagraph{Data.}
We collect 111 videos with verified playback speeds by reviewing online sources that explicitly state the playback rate in the video title or description. 

\myparagraph{Metrics.}
We evaluate the quality of speed predictors using the following metrics: 
(1) Pearson correlation coefficient $\rho$, which measures the linear relationship between predictions and ground truth; (2) Spearman’s rank correlation $r_s$, which evaluates the correctness of their ordinal relationships; and (3) root mean squared error (RMSE).  We transform playback speeds to log space to ensure scale invariance. For interpretability, we also report $e^{\text{RMSE}}$, which reflects the average multiplicative deviation between predictions and ground truth in linear space. 

\myparagraph{Baselines.} We consider four baselines: Gemini 2.5~\cite{comanici2025gemini}, SpeedNet~\cite{benaim2020speednet}, Pulse-of-Motion~\cite{gao2026pulse}, optical flow magnitude, and expert human judgment.
For Gemini, we increase its sampling rate to 8 FPS
to better capture temporal cues
and prompt it to estimate playback speed. For the flow-based baseline, we use SEA-RAFT~\cite{sea-raft} to compute the optical flow between consecutive frames and 
then aggregate the flow magnitude by averaging it across the entire video.
Since flow magnitude is only used as a relative, but not an absolute, measure of speed, we do not report absolute error metrics (\eg, RMSE).
For human evaluation, we design an interface that allows annotators to interactively adjust playback rate until the perceived motion feels like the real-world speed. In practice, it takes on average $\sim$40 seconds to annotate the speed of a video. 

\myparagraph{Results.}
As shown in \cref{tab:speed_prediction}, our method significantly outperforms Gemini, SpeedNet, and the optical-flow baseline, and substantially narrows the gap between human and model performance.

\subsection{Speed-conditioned Video Generation}
\label{sec:experiments-speed-controlled}
\myparagraph{Setup.} 
We set the playback speeds of our model directly via control values of 1, 4, 7, and 10, which corresponds to increasingly faster motion. 
We then utilize 56 image–prompt pairs from the test split of~\dataset, spanning diverse scenes and motion patterns, for evaluation.

\myparagraph{Metrics.}
To measure speed controllability, we compute the average optical flow magnitude for each speed condition and examine whether the resulting magnitudes follow the expected speed ordering.
To assess video generation quality, we compute FID~\cite{fid} and FVD~\cite{fvd} on all 48 text-image–speed triplets from the evaluation set that have corresponding ground-truth videos.

\myparagraph{Baselines.}
We compare against the base model, Wan2.1~\cite{wan2.1}, and a trajectory-based motion control model, ATI~\cite{ati}. Since Wan2.1 does not natively support playback speed control, we approximate different speeds using prompt modifiers 
(\emph{e.g.}, ``ultra slow-motion,'' ``slow-motion,'' ``slightly slow-motion,'' and ``normal''). For the trajectory-based baseline, we first generate a normal-speed video using Wan2.1 and extract its dense tracklets using CoTracker3~\cite{cotracker3}. We then linearly interpolate these tracklets to obtain motion trajectories that are 1/4/16/64 times slower and provide them to ATI as trajectory conditions.

\myparagraph{Results.} 
We report FID and FVD in \cref{tab:i2v-results}, and optical flow analysis in \cref{fig:results}(a).
Our model achieves higher slow-motion generation quality while exhibiting superior speed controllability. As shown qualitatively in \cref{fig:i2v-vis}, videos generated with smaller playback-speed values appear slower than those produced with larger values, indicating that the conditioning signal is effectively learned.

\subsection{Extreme Temporal Super-Resolution}
\label{sec:experiments-super-res}

\myparagraph{Setup.} Evaluation is performed on both DAVIS~\cite{davis} and our~\dataset under three settings: (1) \emph{Clear-input}: on both datasets, we subsample every 8th frame and have the model interpolate frames in between; (2) \emph{Blurred-input}: we synthesize motion blur in low-FPS test videos following the same procedure described in \cref{sec:temporal-super-res}, then have the models reconstruct the original clear \dataset videos; (3) \emph{Real-input}: Since DAVIS videos are normal speed and already contain strong motion blur, we have the models convert them to $8\times$ slow motion, testing their ability to enhance real-world, low-FPS footage. Since ground-truth slow motion is unavailable, we conduct a user study for evaluation.

\myparagraph{Baselines.} We benchmark our method against four interpolation models: FILM~\cite{film}, LDMVFI~\cite{ldmvfi}, Generative Inbetween~\cite{generative-inbetween}, and vanilla Wan2.1-VACE~\cite{jiang2025vace}. For models that generate only a single intermediate frame, we recursively apply them to achieve $8\times$ temporal super-resolution.

\myparagraph{Metrics.} We report standard temporal super-resolution metrics
(\emph{e.g.}, LPIPS~\cite{lpips}, FID~\cite{fid}, FloLPIPS~\cite{flolpips}, and FVD~\cite{fvd})
along with human preference rates.
For the user study, human raters are presented with two videos in each trial, one from our model and the other from a baseline, and asked to select the better one.

\myparagraph{Results.} We show the quantitative comparison under the \emph{Clean-input} setting in \cref{tab:results-woblur}, those under the \emph{Blurred-input} setting in \cref{tab:results-blur}, the user study results across all settings in \cref{fig:results}(b), and qualitative results in \cref{fig:vif-vis}.  Our model consistently outperforms all baselines and achieves over 90\% user preference in the \emph{Real-input} setting (\cref{fig:results}(b)). These results demonstrate our model's superior ability to model real-world dynamics at high temporal resolution.

\begin{table}[t]
    \begin{minipage}[t]{0.48\linewidth}
        \centering
            \centering\small
            \begin{tabular}{l|cccc}
            \toprule
            Method & ~IP~ & $\rho$ $\uparrow$ & $r_s$ $\uparrow$ & RMSE $\downarrow$ \\
            \midrule
            \multirow{2}{*}{VideoLLM~\cite{comanici2025gemini}~}  & & 0.426        & 0.308        & 1.568     \\
             & \checkmark & \textbf{0.552}   & \textbf{0.479}        & \textbf{1.221} \\
            \midrule
            \multirow{2}{*}{Ours} &   & 0.680        & 0.684        & 0.917 	 \\
            & \checkmark    & \textbf{0.735}   & \textbf{0.706}   & \textbf{0.649} \\
            \bottomrule
        \end{tabular}
        \caption{{\bf Ablation on iterative prediction (IP).} We conduct an ablation study on the IP mechanism. The results show that the inclusion of IP consistently improves performance for both VideoLLM and our self-supervised model, demonstrating its robustness in enhancing speed prediction accuracy.
        }    
        \label{tab:iterative_prediction}
    \end{minipage}
    \hfill
    \begin{minipage}[t]{0.48\linewidth}
            \centering\small
            \begin{tabular}{l|cccc}
            \toprule
            Training data & ~IP~ & $\rho$ $\uparrow$ & $r_s$ $\uparrow$ & RMSE $\downarrow$ \\
            \midrule
            \multirow{2}{*}{Standard Videos~} & & 0.242        & 0.223        & 1.232     \\
             & \checkmark & 0.632   & 0.598        & 0.737 \\
            \midrule
            \multirow{2}{*}{\dataset}  & & 0.680        & 0.684        & 0.917 	 \\
              & \checkmark & \textbf{0.735}   & \textbf{0.706}   & \textbf{0.649} \\ 
            \bottomrule
            \end{tabular}
        \caption{{\bf Playback-speed prediction using different training data.} We train the speed predictor using either \dataset or only standard datasets (Adobe240fps~\cite{adobe240} and normal-speed videos). The results show that exposing the model to a broad range of temporal scales improves its speed prediction capability.
        }        
        \label{tab:speed-prediction-dataset}
    \end{minipage}
\end{table}

\subsection{Analysis}
\label{sec:experiments-dataset}

\begin{wrapfigure}{r}{0.485\textwidth}
    \centering
    \IfDefinedSwitch{\embedVideo}{
    \animategraphics[autoplay,loop,width=\linewidth, trim=0 0cm 0 0cm, clip]{16}{figures/i2v-dataset/}{001}{080}
    }{
        \includegraphics[width=1.\linewidth]{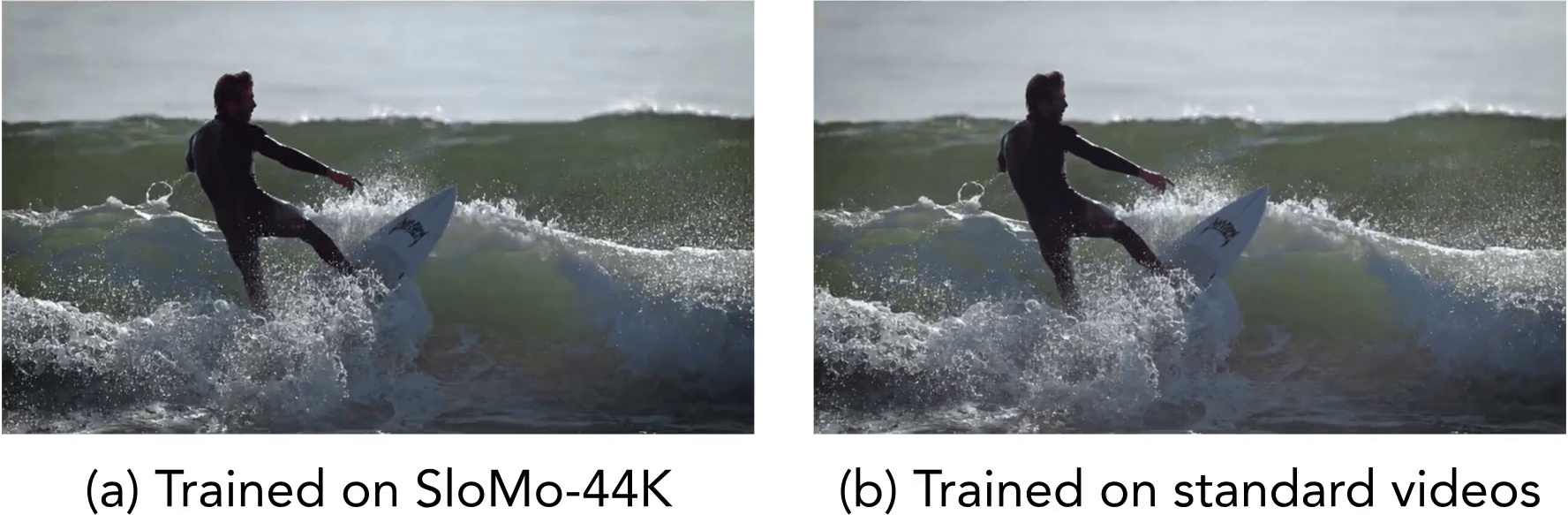}
    }
    \vspace{-5mm}
    \caption{
        \textbf{Speed-controlled video generation with different training data.} Training on standard-FPS videos with artificial slowdowns leads to stuttering artifacts, while training on \dataset results in realistic slow-motion dynamics. \it{Videos are embedded. Please view with Adobe Acrobat Reader.}
    }
    \label{fig:i2v-dataset}
\end{wrapfigure}

In this section, we conduct ablation studies to examine the impact of iterative prediction on speed estimation and to assess the importance of using \dataset to train both speed predictors and speed-conditioned video generation models.
    
\myparagraph{Importance of Iterative prediction.}
The ablation study in \cref{tab:iterative_prediction} shows that IP enhances the accuracy of both our model and VideoLLM, confirming the technique's robustness across models.

\myparagraph{Importance of \dataset.}
Is \dataset necessary for training our proposed models? To answer this question, we first compare our speed predictor (trained on \dataset) with a baseline trained solely on standard datasets (Adobe240fps~\cite{adobe240} and normal-speed videos). The result in \cref{tab:speed-prediction-dataset} shows that models exposed to a wider variety of speeds outperform those trained only on standard videos. We further evaluate our speed-conditioned model against a baseline trained on videos artificially slowed from standard-FPS footage, following~\cite{huang2025spacetimepilot}.
As shown in \cref{tab:i2v-dataset,fig:i2v-dataset}, the baseline exhibits stuttering artifacts, whereas our model faithfully synthesizes slow-motion dynamics.
\section{Discussion}

In this work, we showcase how we can leverage the multimodal cues, temporal structure inherently present in videos, and the vast amount of slow motion videos available on the internet to develop models that can perceive and manipulate the flow of time.
While we achieve state-of-the-art performance across both understanding and generation tasks, we still encounter several limitations. For instance, our speed understanding models can be misled when videos contain limited motion cues or when people deliberately move slowly; our generation models also rely on a pretrained Wan backbone, leaving room for improvement through architectural innovations or full fine-tuning.
We believe this work establishes time as a manipulable dimension in video learning and opens new directions for temporal forensics and richer world models that capture how events unfold.

\section*{Acknowledgment}
The research is partially supported by a gift from Ai2, NVIDIA Academic Grant, DARPA TIAMAT program No. HR00112490422, and the National Science and
Technology Council via grants NSTC 114-2634-F-002-006 and NSTC 114-2640-E-002-006. Its contents are solely the responsibility of the authors and do not necessarily represent the official views of DARPA.
We also thank Xuyi Meng, Chuanruo Ning, and Guangzhao He for their valuable contributions and discussions, and the National Center for High-performance Computing (NCHC) for providing computational and storage resources.

\section*{Author Contributions}
\textbf{Yen-Siang Wu} led the project, designed and curated the \dataset dataset, proposed the core methodology, and conducted the primary experiments and manuscript drafting. \textbf{Rundong Luo} helped data curation, contributed to the design, training, and evaluation of the video generation models, performed the literature review and manuscript drafting. \textbf{Jingsen Zhu} executed the preliminary experiments that informed the development of the proposed approach. \textbf{Tao Tu} managed the user studies, prepared the visual demonstrations, and assisted in proofreading.


\bibliographystyle{plainnat}
\bibliography{main}

\clearpage
\appendix

\section{Supplementary Material Overview}

In this supplementary document, we provide additional implementation details and present extended qualitative results. For video demonstrations, please visit the \href{https://seeing-fast-and-slow.github.io/}{project page}.

\begin{figure}[h]
    \centering
    \includegraphics[width=\linewidth]{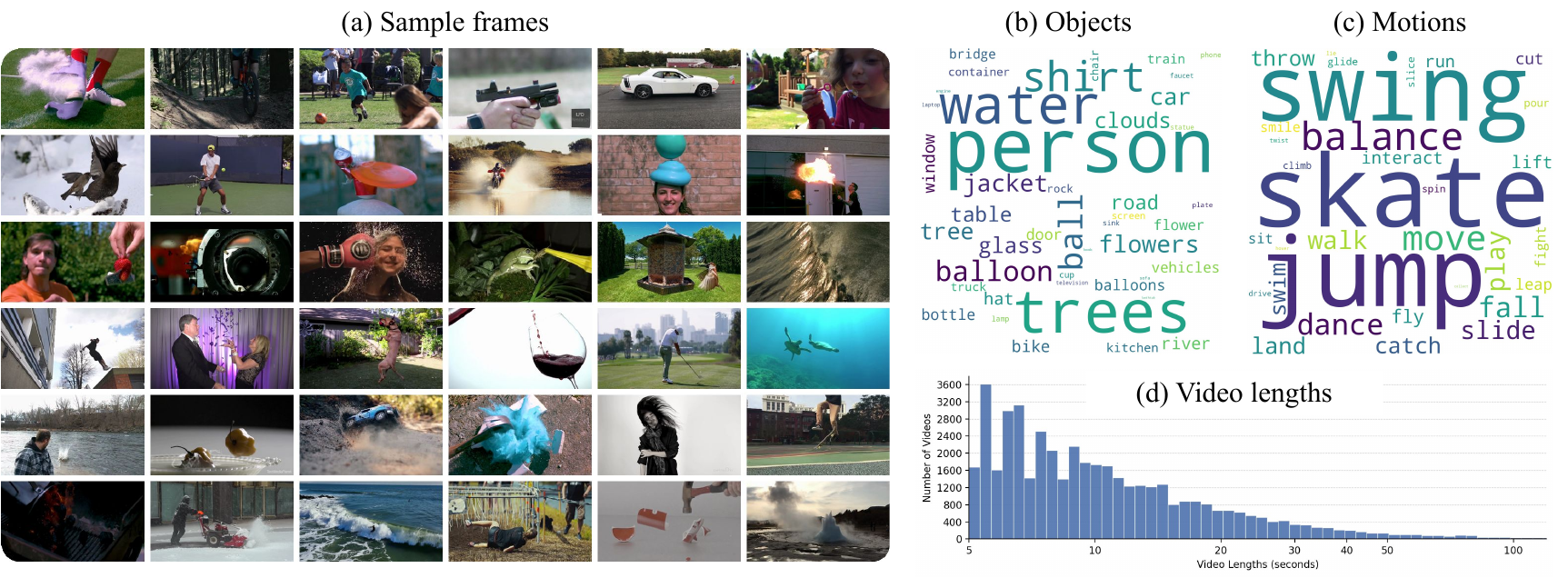}
    \vspace{-6mm}
    \caption
    {\textbf{Dataset overview.}
    (a) Sample frames illustrating the dataset's contextual diversity;
    (b) Word cloud of nouns from generated captions, capturing a broad range of objects and scenes;
    (c) Word cloud of verbs, reflecting diverse human and object motions;
    (d) Histogram of video lengths, showing variation in temporal scales.}
    \label{fig:dataset_overview}
\end{figure}

\section{Dataset Details}
\label{sec:dataset-details}
In this section, we provide additional details on the curation and statistical composition of the \dataset dataset.

\subsection{Dataset Curation}

We expand upon the data curation pipeline introduced in Sec.~3.3 of the main paper.

\myparagraph{Video Acquisition and Generic Filtering.} 
We collect videos from YouTube, Vimeo, and Flickr using queries such as ``high frame rate,'' ``high-speed camera,'' and ``slow motion,'' along with related synonyms. 
Following \cite{blattmann2023stable,luo2025beyond}, we apply TransNetv2~\cite{soucek2024transnet} for shot segmentation and use an OCR model~\cite{ye2023dptext} to remove clips with excessive text overlays. 
To further ensure quality, we filter out CGI and screen-captured content using Qwen2.5-VL~\cite{bai2025qwen2}, and discard low-quality videos based on VQA metrics~\cite{he2024cover}.

\myparagraph{Slow-Motion–Focused Processing.} 
Because many online videos exhibit heterogeneous or unspecified playback speeds, we employ our speed-change detector (main paper Sec.~3.1) to segment each video into clips with homogeneous playback rates.  
However, these segments are not necessarily slow motion; without additional filtering, the dataset would remain dominated by standard $1.0\times$ content.
We therefore introduce a dedicated slow-motion video detection approach to identify and retain only clips that exhibit reliably slow-motion characteristics. See the next section for implementation details.

\myparagraph{Annotation.} Finally, we annotate each clip with a predicted playback speed using our speed estimator. 
We also generate dense captions—including short and long descriptions, and attributes such as background, style, shot type, lighting, and atmosphere, to capture both semantic and aesthetic aspects of the scene, using InternVL3~\cite{zhu2025internvl3}. 

\subsection{Slow-Motion Video Classification}
\label{sec:filtering-slow-motion-content}

As discussed above, clips extracted from raw videos are not guaranteed to contain slow motion. Without further filtering, the dataset would be dominated by normal-speed content. We therefore develop a dedicated slow-motion filtering pipeline to identify genuine slow-motion clips.

\myparagraph{Approach.} To filter slow-motion videos, we adopt two complementary strategies:
(1) We query a VideoLLM with each original video (as downloaded online) and prompt it to localize slow-motion segments.
(2) We train a ViT-based video classifier on a small set of human annotations to predict whether a 2-second clip is in slow motion.
Because VideoLLMs leverage global semantics and contextual cues, while ViT-based classifiers specialize in fine-grained visual patterns, we combine both predictions in a two-stage filtering pipeline.

\myparagraph{Implementation Details.}
In practice, we use Gemini~2.5 as the VideoLLM and a fine-tuned VideoMAEv2 as the clip-level classifier. To obtain training data, eight annotators label 2{,}400 clips as \texttt{slow motion}, \texttt{not slow motion}, or \texttt{unknown}, using all available information including video titles and descriptions. Clips labeled \texttt{unknown} are discarded, and the remaining ones are used for training and evaluation.

\myparagraph{Results.}
On our curated validation set, Gemini achieves 74.8\% accuracy, while the finetuned classifier reaches 84.4\%. To leverage the complementary strengths of both models, we combined their predictions. We discard clips if either: (1) Gemini identifies less than 10\% of its content as slow motion, or (2) the video classifier assigns a slow-motion probability below 0.998. This combined filtering achieves 98\% precision and 44\% recall in retaining slow-motion clips.

\subsection{Dataset Statistics}
\label{sec:dataset-statistics}

\begin{figure}[t]
    \centering
    \includegraphics[width=.5\linewidth]{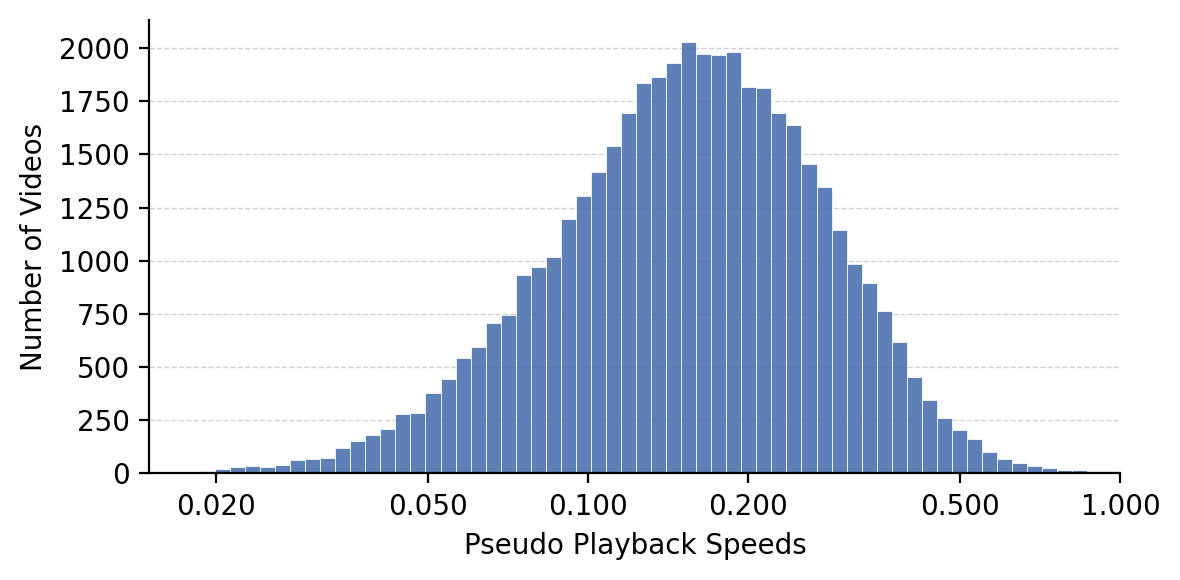}
    \caption{
    \textbf{Histogram of pseudo-speed annotations.} \dataset spans a wide spectrum of playback speeds.
    }
    \label{fig:video_speeds}
\end{figure}

\cref{fig:dataset_overview} presents an overview of \dataset, including sample frames, two word clouds of nouns and verbs extracted from the generated video captions, and a histogram of video lengths. \cref{fig:video_speeds} shows the histogram of pseudo-speed annotations. As shown, the dataset covers a diver range of scenes, objects, motions, lengths, and temporal scales, making it well-suited for learning all sorts of real-world motions and training generic video generation models.

\section{Experimental Details}
\label{sec:experimental-details}

In this section, we provide additional experimental details, including more qualitative results, ablation studies, and information on the user studies.

\subsection{Speed Estimation}
\label{sec:perceiving-the-flow-of-time-details}

\noindent\textbf{Equivariance in Speed Estimation.} A key concept underlying our playback speed estimator is \emph{equivariance}, which we elaborate on here.

Consider an ideal estimator $f_\theta$ that perfectly predicts the playback speed of any video. For a video $\mathbf{V}$ with playback speed $x_\mathbf{V}$, accelerating it by a factor of $k$ (via temporal subsampling) yields a new video whose effective playback speed becomes $k\cdot x_\mathbf{V}$. 
Let ``$\cdot $'' denote temporal acceleration so that $k\cdot \mathbf{V}$ represents the $k$-times–sped-up version of $\mathbf{V}$. 
Then a perfect speed estimator should satisfy:
\begin{equation}
    f_\theta(k\cdot\mathbf{V}) = k \cdot f_\theta(\mathbf{V}),
    \label{eq:equivariance}
\end{equation}
which mirrors the definition of \emph{equivariance}. We refer to this property as the \emph{equivariance of speed estimation} accordingly.

\begin{figure}[t]
    \centering
    \includegraphics[width=\linewidth]{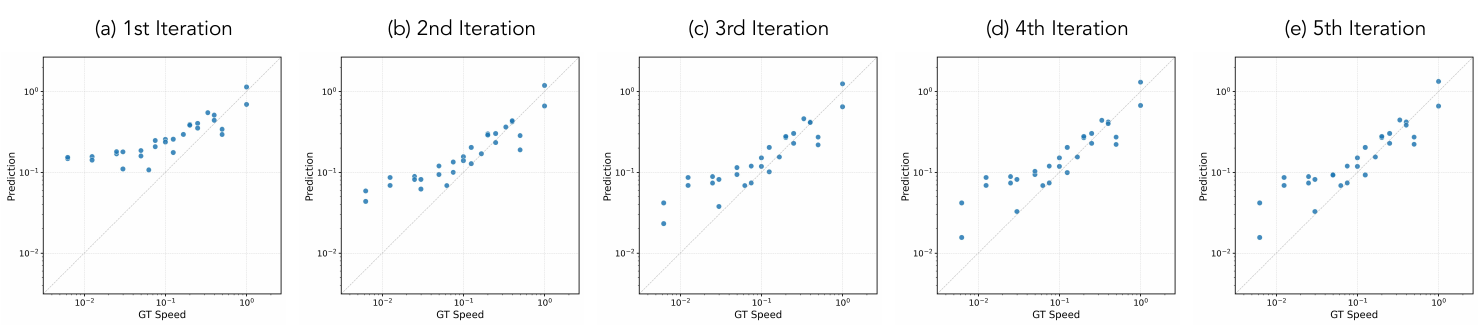}
    \vspace{-4mm}
    \caption{
    \textbf{Ablation on iterative prediction.} We vary the number of iterations and show the corresponding speed prediction. As shown, performance improves notably by the third iteration, while additional iterations yield minimal gains, showing the predictions have converged.
    }
    \vspace{-3mm}
    \label{fig:speed-ablation}
\end{figure}

\myparagraph{Ablation Study: Iterative Prediction.} We study how iterative prediction affect speed estimates in \cref{fig:speed-ablation}. The first iteration produces relatively inaccurate predictions, particularly for extremely slow videos. By the third iteration, predictions are substantially improved, demonstrating the benefit of iterative prediction. Additional iterations (up to five) result in negligible changes, indicating the predictions have converged. Based on these findings, we adopt three iterations in practice to balance accuracy and computation.

\begin{figure}[p]
    \centering
    \includegraphics[width=0.49\linewidth, trim=0.0cm 0cm 0.0cm 0cm, clip]{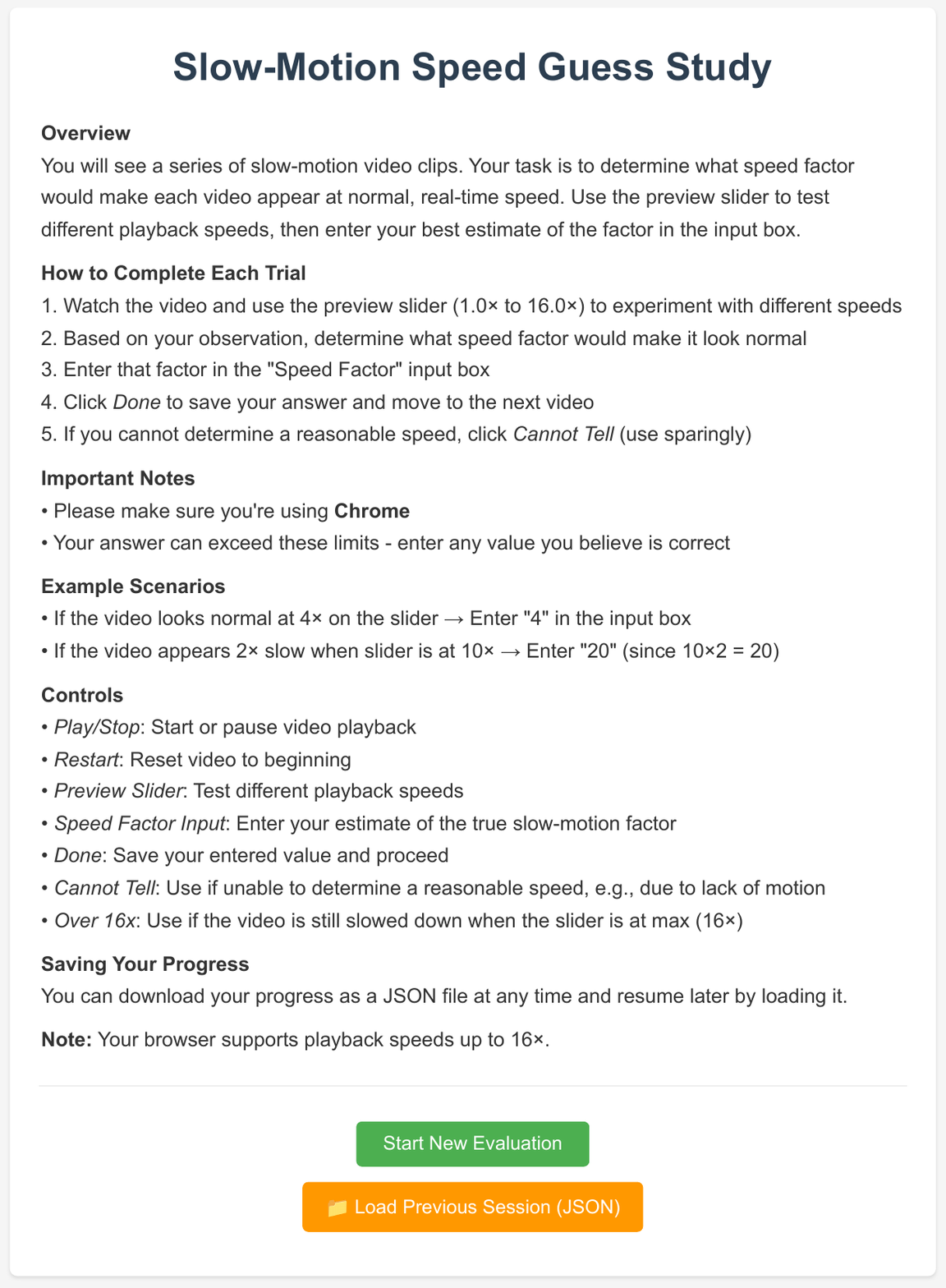}
    \hspace{0.00\linewidth}
    \includegraphics[width=0.49\linewidth, trim=0.0cm 0cm 0.0cm 0cm, clip]{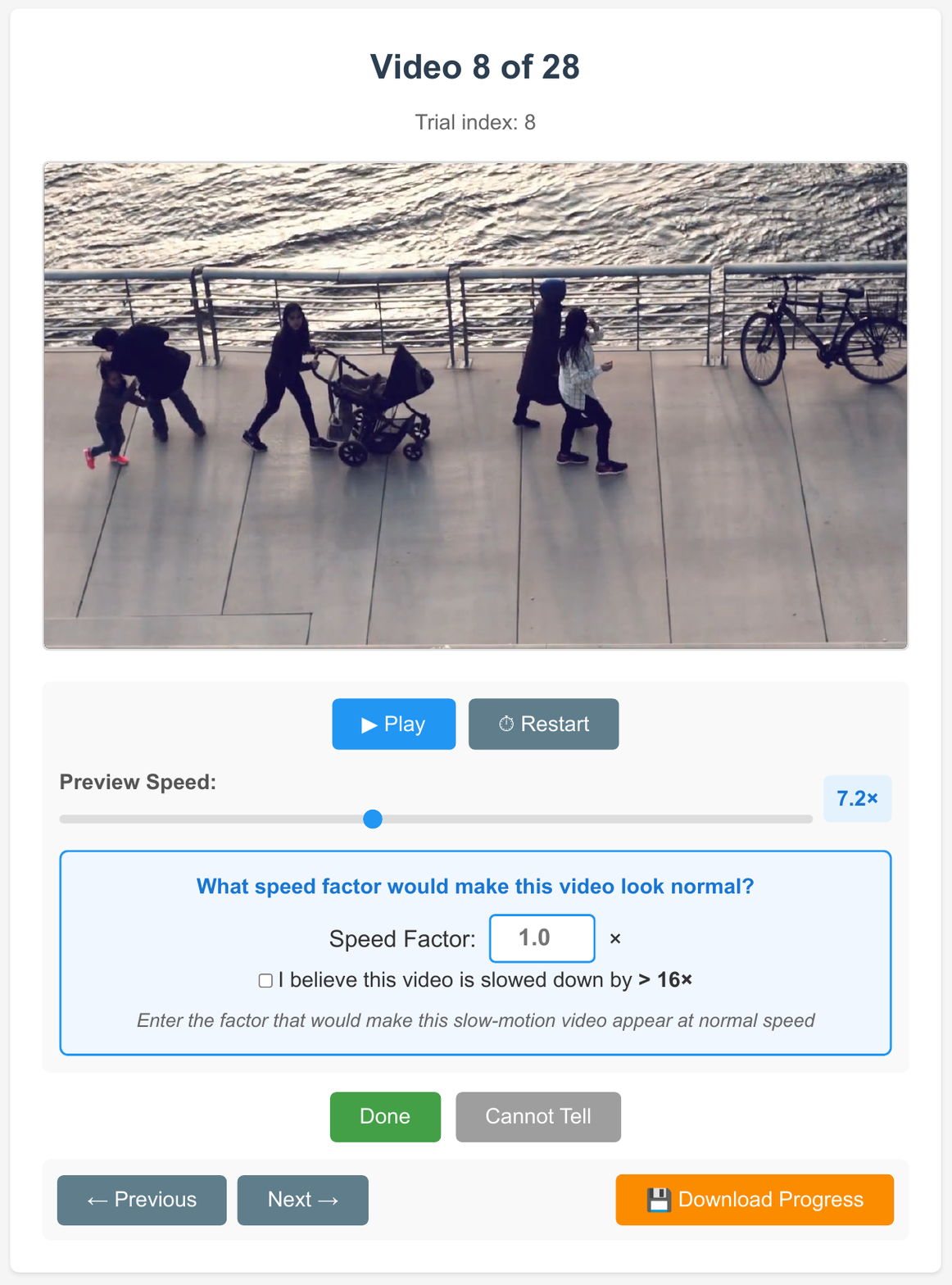}
    \caption{
    \textbf{Human Annotation interface for video speed prediction.} The interface provides a slider that allows annotators to accelerate videos until the motion appears aligned with its real-world progression speed. The final speed estimate is then computed from the selected speed-up factor. This design ensures that human perceptual sensitivity is used to its fullest.
    }
    \label{fig:speed-prediction-screenshot}
\end{figure}

\myparagraph{Human Annotation Interface.}
We show a screenshot of our human annotation interface in \cref{fig:speed-prediction-screenshot}. 
The interface is designed to fully leverage human perceptual sensitivity when judging playback speed.

\subsection{Speed-conditioned Video Generation}

\noindent\textbf{Training Details.} To prevent any speed bucket from dominating the training distribution, we balance the sampling probability across all buckets. We finetune Wan2.1-I2V-14B-480P~\cite{wan2.1} on 4~GB200 GPUs for two days, using a learning rate of $1\times10^{-5}$ for the speed-conditioning modules and $1\times10^{-4}$ for the LoRA adapters.

\myparagraph{Evaluation Details.} As discussed in Sec. 5.3., we generate 224 videos per baseline during evaluation (56 text-image pairs, each evaluated at four different speed settings). However, when evaluating slow-motion generation quality, we cannot compute FID and FVD across all text–image–speed combinations, since each text-image pair in the evaluation set is associated with only a single speed. In fact, only 48 image–text–speed triplets in the evaluation set have corresponding ground-truth videos at matching speeds. The remaining triplets either contain too few frames after speeding up to the target speed or have frame rates that are too low to slow down to the desired speed. Consequently, we use this subset of 48 triplets to compute FID and FVD scores for overall generation quality.

For completeness, we also report VBench~\cite{huang2024vbench} results here. Nonetheless, VBench does not accurately capture slow-motion generation quality, as it (1) is designed to evaluate \emph{standard-speed} videos and (2) primarily assesses whether the video appears visually convincing instead of whether it follows real-world principles~\cite{zheng2025vbench2}.

\begin{figure}[p]
    \centering
    \includegraphics[width=\linewidth]{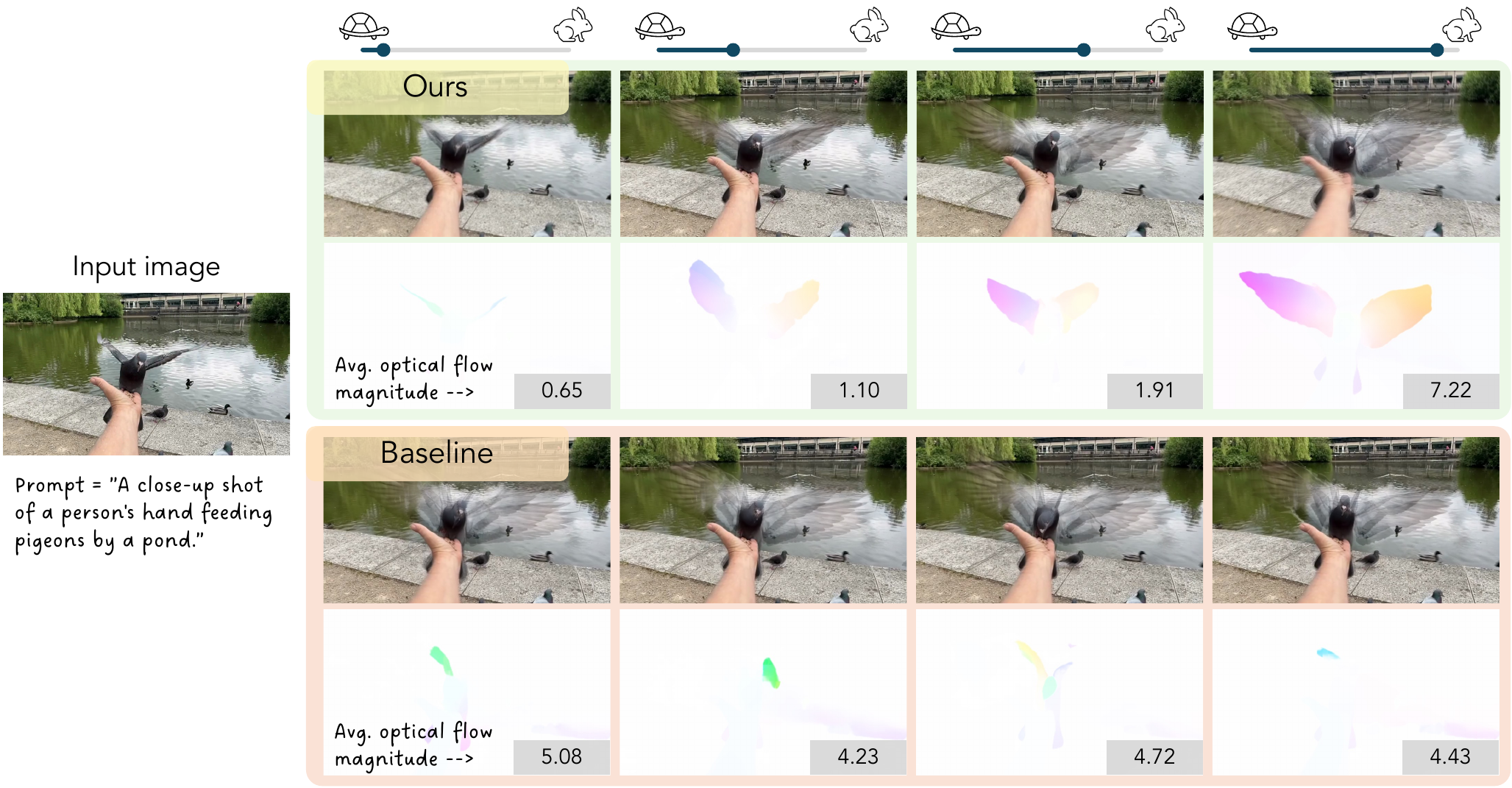}
    \caption{
    \textbf{Qualitative comparisons for speed-conditioned video generation.} 
    In each group, the first row shows the average image of the generated video, and the second row visualizes the optical flow between the first and fifth frames. 
    For our method, slower speed conditions yield crisp average images (as the scene undergoes minimal change) and low-magnitude optical flow (faint colors). 
    As the conditioned playback speed increases, the average image becomes progressively blurrier and the optical flow magnitude increases (more saturated colors). 
    In contrast, the baseline model produces videos with nearly unchanged temporal dynamics across different speed conditions.}
    \label{fig:more-speed-control}
\end{figure}

\myparagraph{Results.} We provide additional qualitative comparisons between our model and baseline methods in \cref{fig:more-speed-control} and additional VBench results in \cref{tab:vbench}. As shown, our approach produces high-quality videos with clear and consistent speed controllability: videos generated with smaller playback-speed values exhibit noticeably slower motion, whereas larger values yield faster dynamics. In contrast, baseline methods show minimal variation across different speed-control prompt modifiers, indicating limited control over temporal dynamics.

\subsection{Temporal Super-Resolution}

\noindent\textbf{Training Details.} Similar to speed-conditioned video generation, we balance the sampling probability across all buckets during training. We attach LoRA adapters to Wan2.1-VACE-14B~\cite{wan2.1} and finetune them for two days on 4 GB200 GPUs, using a learning rate of $1\times 10^{-4}$. We train one model for the traditional \emph{Clear-input} setting and one model for the \emph{Blurred-input} or \emph{Real-input} setting, where low-FPS inputs exhibit stronger motion blur.

\begin{figure}[p]
    \centering
    \includegraphics[width=0.47\linewidth, trim=1.0cm 0cm 1.0cm 0cm, clip]{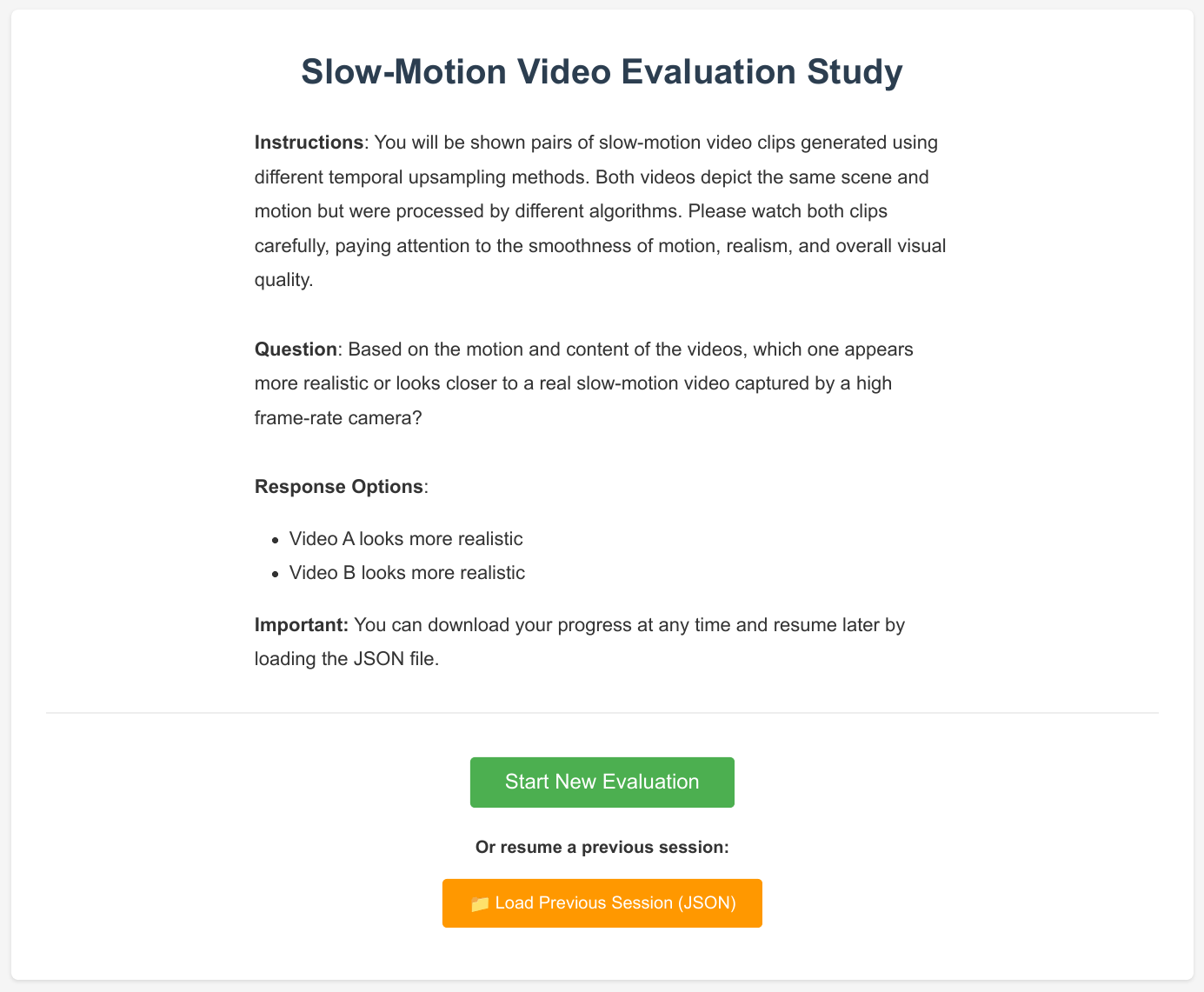}
    \hspace{0.02\linewidth}
    \includegraphics[width=0.47\linewidth, trim=1.0cm 0cm 1.0cm 0cm, clip]{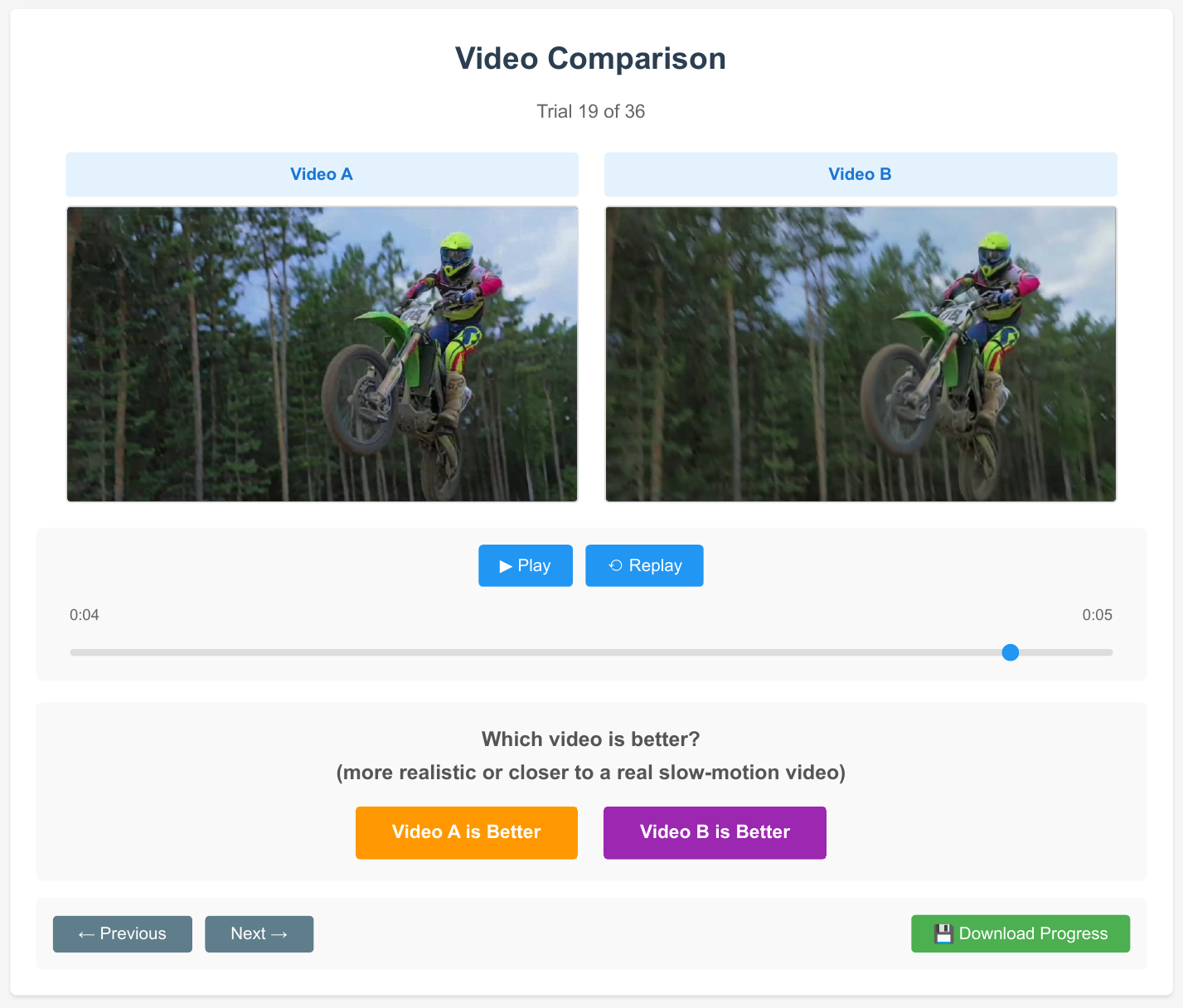}
    \caption{
    \textbf{User study interface for temporal super-resolution.} Participants are instructed to compare playback speeds based on motion smoothness and temporal details. In each trial, two videos are presented side-by-side and the participants are asked to select the one that appears slower.
    }
    \label{fig:tsr-screenshot}
\end{figure}

\begin{figure}[p]
    \centering
    \includegraphics[width=\linewidth]{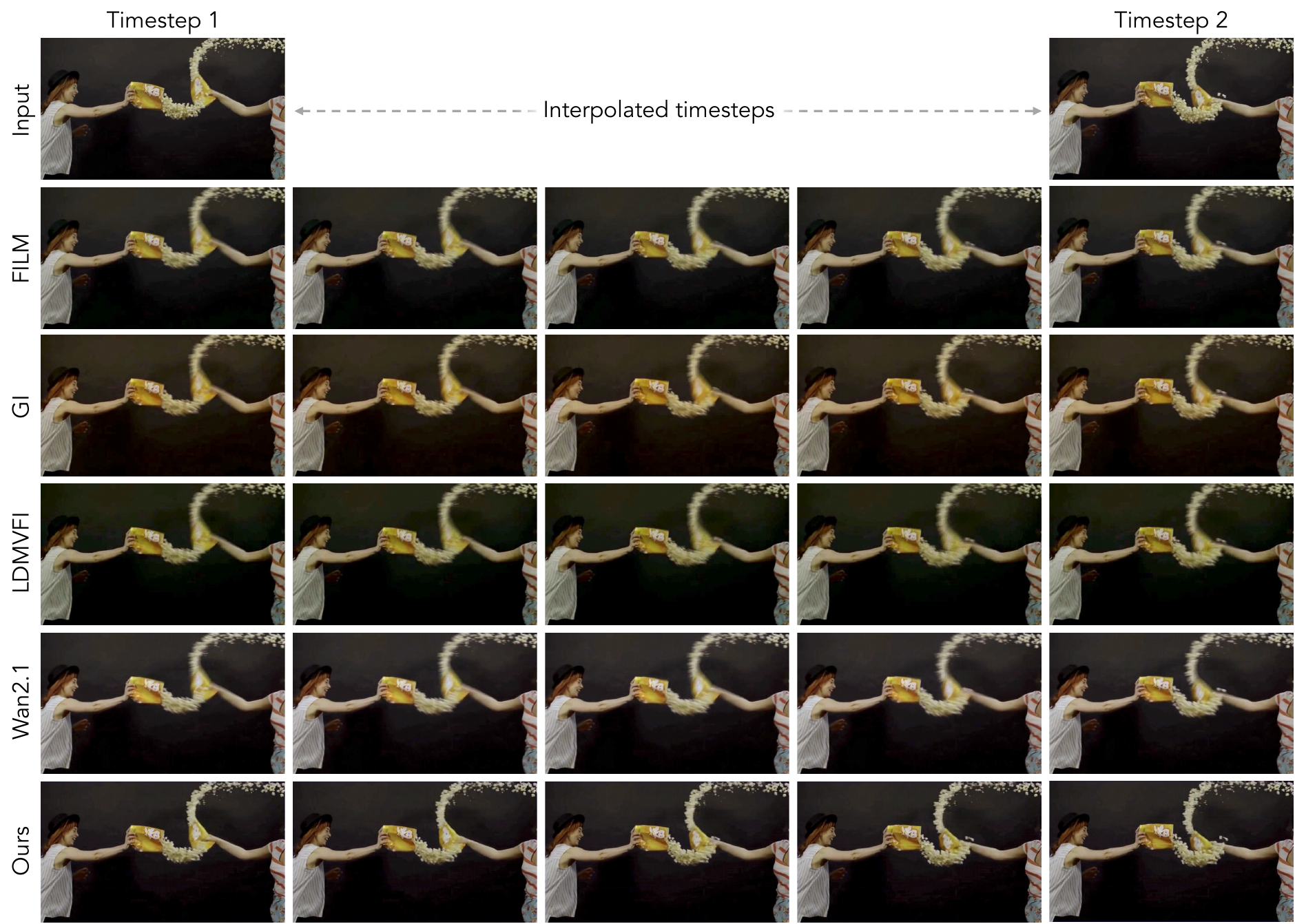}
    \caption{
    \textbf{Qualitative comparisons for temporal super-resolution on \dataset\ (blurred-input).}
    Given two input frames, each method interpolates multiple timesteps. Prior methods produce over-blurred popcorn trajectories and inconsistent boundaries, whereas our method yields smoother motion and sharper, more coherent details under strong motion blur.}
    \label{fig:tsr_ours_blurred}
\end{figure}

\begin{figure}[p]
    \centering
    \includegraphics[width=\linewidth]{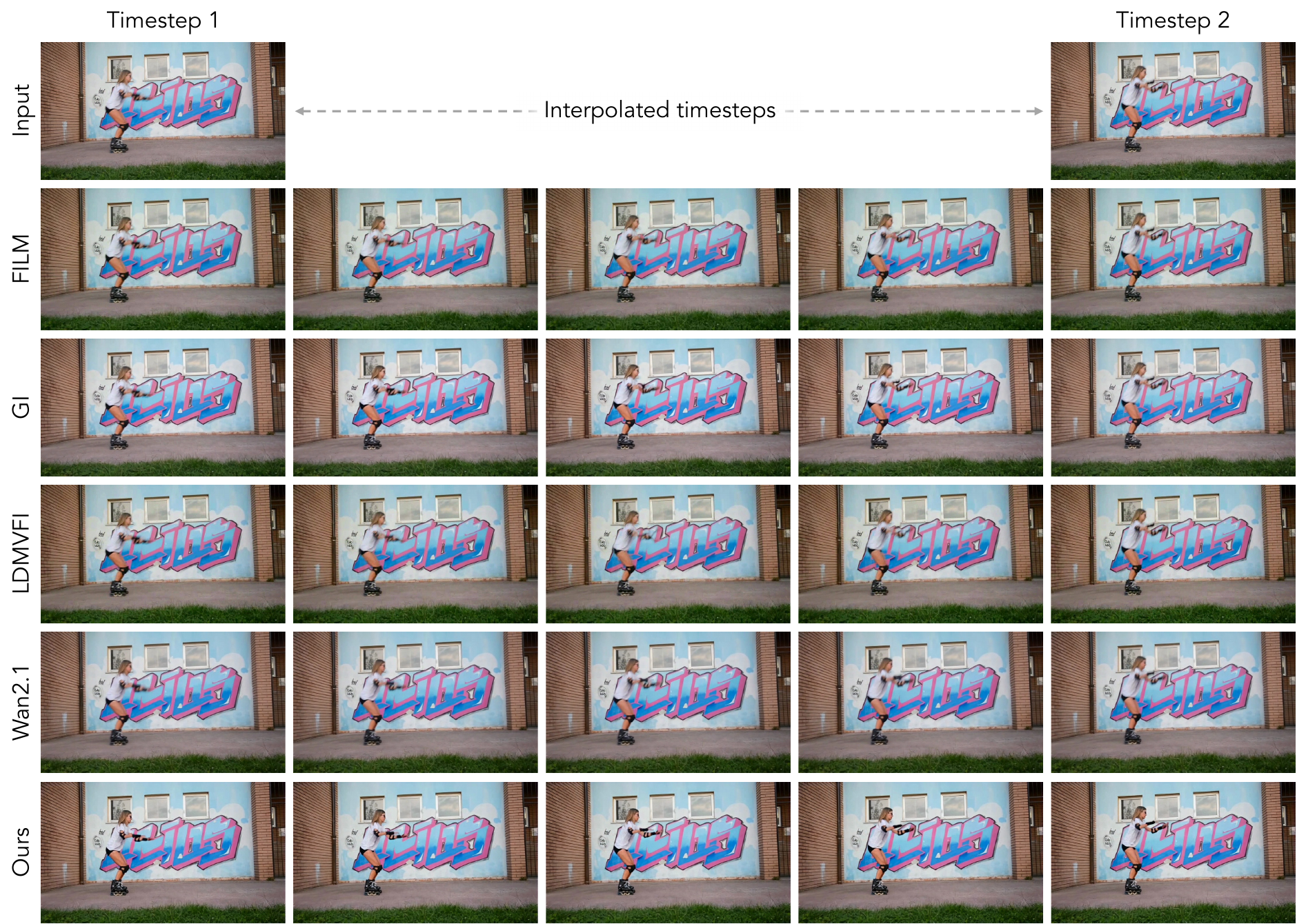}
    \caption{
    \textbf{Qualitative comparison for temporal super-resolution on DAVIS (real-input).} 
    Our method produces smoother motion trajectories and sharper fine-grained details than traditional temporal super-resolution approaches (e.g., see the girl’s arms). These results highlight the benefit of explicitly synthesizing motion blur for low-FPS inputs and training on our curated slow-motion dataset.}
    \label{fig:tsr_davis_real}
\end{figure}

\myparagraph{Results.} We present qualitative comparisons between our method and all baselines under the \emph{Blurred-input} setting in Fig.~\ref{fig:tsr_ours_blurred} and the \emph{Real-input} setting in Fig.~\ref{fig:tsr_davis_real}. As shown, our approach generates clear and temporally consistent intermediate frames even from heavily blurred inputs, outperforming traditional temporal super-resolution methods. We also provide the user study interface for temporal super-resolution in \cref{fig:tsr-screenshot}. For additional qualitative results, please refer to our \textbf{project page} in the supplementary material.

\begin{table}[t]
    \centering\footnotesize
    \begin{tabular}{lcccc}
    \toprule
    Method & \makecell{Aes.\\quality}$\uparrow$ & \makecell{Imag.\\quality}$\uparrow$ & \makecell{Motion\\smooth.}$\uparrow$ & \makecell{Temp.\\Flicker.}$\uparrow$\\ \midrule
    Wan2.1 & 0.5384 &	\textbf{0.6367} &	0.9803 &	0.9638 \\
    \textbf{Ours} & \textbf{0.5394} &	0.6337 &\textbf{0.9871} &	\textbf{0.9700} \\ \bottomrule
    \end{tabular}
    \caption{\textbf{VBench results for speed-controlled video generation.} For completeness, we compare Wan2.1 with our speed-controlled video generation model on VBench metrics. Notably, VBench does not reliably reflect the quality of slow-motion generation because it (1) is designed to evaluate \emph{standard-speed} videos and (2) primarily assesses whether the video appears visually convincing instead of whether it follows real-world principles~\cite{zheng2025vbench2}.}
    \label{tab:vbench}
\end{table}

\section{Comparison to BulletTime and SpaceTimePilot}

Recent concurrent works, BulletTime~\cite{wang2025bullettime} and SpaceTimePilot~\cite{huang2025spacetimepilot}, also explore time-controlled video generation. However, they differ from our work in several fundamental ways. First, these work focus on \emph{video editing}, enabling users to manipulate each frame's timestamp through a \emph{relative} time-remapping. In contrast, our speed-conditioned framework addresses \emph{image-to-video generation} conditioned on \emph{absolute} temporal speed, which requires the model to have knowledge of real-world motion speeds. In other words, BulletTime and SpaceTimePilot modify \emph{existing} motion on a \emph{relative} time scale, while our approach synthesizes motion \emph{from scratch} according to \emph{absolute} speed conditions.

Moreover, our speed-conditioned model is capable of modeling complex dynamics across a broad range of temporal granularities (e.g., up to 100$\times$ slow motion), as it is trained on a real-world high-frame-rate video dataset, \dataset. In contrast, BulletTime and SpaceTimePilot are finetuned only on a synthetic video dataset, which limits its ability to model intricate real-world motions at high frame rates (\eg, rapid wing flapping), and primarily supports simple slow-motion effects, such as human body movements.

Importantly, beyond time-controlled video generation, our work also investigates speed perception methods, enabling applications in video forensics and leading to the construction of the \dataset dataset. 


\end{document}